\pdfoutput=1

\documentclass[11pt]{article}

\usepackage[final]{coling}

\usepackage{times}
\usepackage{latexsym}

\usepackage[T1]{fontenc}

\usepackage[utf8]{inputenc}

\usepackage{microtype}

\usepackage{inconsolata}

\usepackage{graphicx}

\usepackage{amsmath}
\usepackage{amssymb}
\usepackage{array}
\usepackage{booktabs}
\usepackage{makecell}
\usepackage{multirow}
\usepackage{colortbl}
\usepackage{xcolor}
\usepackage{adjustbox}
\usepackage{lscape}
\usepackage{pdflscape}
\usepackage{rotating}
\usepackage{caption}
\usepackage{float}
\usepackage{siunitx}
\usepackage{lipsum} 
\usepackage{newfloat}
\usepackage{listings}
\usepackage{inconsolata}
\usepackage{booktabs}
\usepackage{xcolor}
\usepackage{graphicx}
\usepackage{subfigure}
\usepackage{adjustbox}
\usepackage{amsmath}
\usepackage{pifont}
\usepackage{diagbox}
\usepackage{multirow}
\usepackage{amsfonts}

\title{\zigzagkv: Dynamic KV Cache Compression for Long-context Modeling 

based on Layer Uncertainty}

\author{
    Meizhi Zhong\textsuperscript{\rm 1}\thanks{Work during Xiaohongshu internship.},\space\space
    Xikai Liu\textsuperscript{\rm 2},\space\space 
    Chen Zhang\textsuperscript{\rm 2},\space\space
    Yikun Lei\textsuperscript{\rm 2}, \space\space
    Yan Gao\textsuperscript{\rm 2},\space\space 
    Yao Hu\textsuperscript{\rm 2},\space\space \\
    \textbf{Kehai Chen}\textsuperscript{\rm 1} \thanks{Corresponding authors},\space\space 
    \textbf{Min Zhang}\textsuperscript{\rm 1}\space\space \\
    \textsuperscript{\rm 1}Institute of Computing and Intelligence, Harbin Institute of Technology, Shenzhen, China \\
    \textsuperscript{\rm 2} Xiaohongshu Inc.\\
    {\tt meizhi.zhong.1999@gmail.com},\space\space
    {\tt chenzhang9702@outlook.com},\space\space \\
    {\tt \{chenkehai,zhangmin2021\}@hit.edu.cn},\space\space \\
    {\tt \{xikai,zhizhu,yadun,xiahou\}@xiaohongshu.com}
}

\newcommand{\fullkv}{\texttt{FullKV}}
\newcommand{\snapkv}{\texttt{SnapKV}}
\newcommand{\ho}{\texttt{H2O}}
\newcommand{\streamingllm}{\texttt{StreamLM}}
\newcommand{\pyramidkv}{\texttt{PyramidKV}}
\newcommand{\zigzagkv}{\texttt{ZigZagKV}}

\begin{document}

\maketitle

\begin{abstract}
Large Language models (LLMs) have become a research hotspot. To accelerate the inference of LLMs, storing computed caches in memory has become the standard technique. 
However, as the inference length increases, growing KV caches might lead to out-of-memory issues.
Many existing methods address this issue through KV cache compression, primarily by preserving key tokens throughout all layers to reduce information loss. Most of them allocate a uniform budget size for each layer to retain.
However, we observe that the minimum budget sizes needed to retain essential information vary across layers and models based on the perspectives of attention and hidden state output.
Building on this observation, this paper proposes a simple yet effective KV cache compression method that leverages layer uncertainty to allocate budget size for each layer.
Experimental results show that the proposed method can reduce memory usage of the KV caches to only $\sim$20\% when compared to Full KV inference while achieving nearly lossless performance.
\end{abstract}

\section{Introduction}

Large language models (LLMs) \citep{radford2018improving,touvron2023llama,zhang2023towards,brown2020language,huang2024multi} have been employed across a wide range of natural language processing tasks, including code completion \citep{rozière2023code} and question answering \citep{kamalloo2023evaluating,jiang2021can,su2019generalizing}. To accelerate inference, it is common practice to store precomputed key and value hidden states in memory as a KV cache. However, as input lengths increase during long-context modeling~\citep{peng2023yarn,ntk,fu2024data,zhong2024understanding,zhang2024modification}, the size of the KV caches grows proportionally, leading to memory consumption and out-of-memory issues. For instance, maintaining a KV caches for 100K tokens in GPU memory for the LLaMA-2 7B model requires over 50GB of memory, compared to less than 1GB for a 2K context.

\begin{figure}[t]
	\centering
	\includegraphics[width=\linewidth]{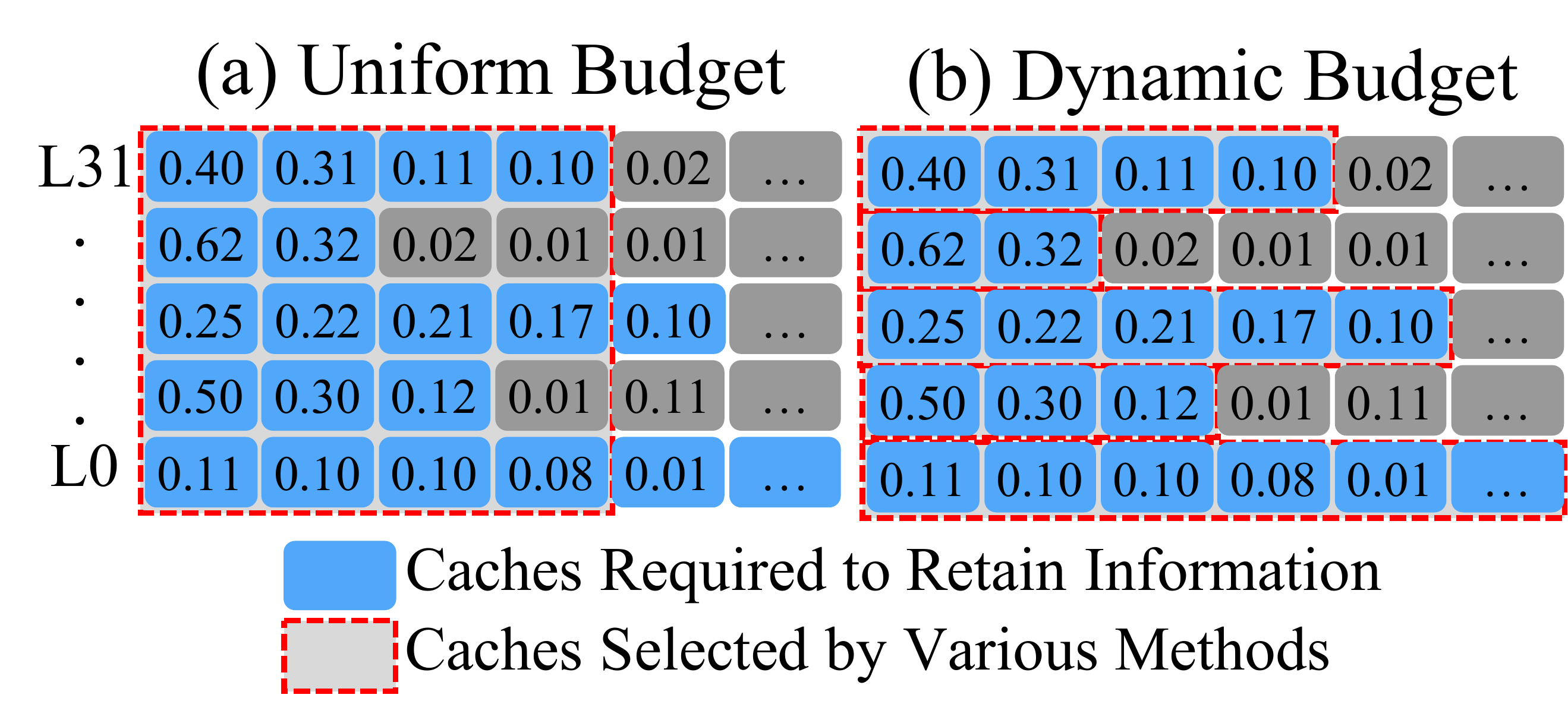} 
	\caption{Comparison of the proposed method (\zigzagkv, Right) with previous PartialKV inference methods (Left). Most existing PartialKV methods allocate a uniform cache size per layer, whereas \zigzagkv~ dynamically adjusts the cache size based on layer uncertainty. ``L0" and ``L31" refer to Layer 0 and Layer 31, respectively. The numbers in brackets represent importance scores from the current token to prefix tokens, sorted from high to low.}
	\label{fig:introduction}
\end{figure}
A straightforward solution to mitigate these memory issues is to reduce the size of the KV caches, thereby decreasing memory usage~\citep{xiao2023efficient,liu2024scissorhands,li2024snapkv,chen2024nacl,ren2024efficacy,zhang2024pyramidkv,yang2024pyramidinfer}. 
The key to these methods lies in evicting nonessential KV caches while minimizing information loss caused by the eviction process.
As depicted in Figure~\ref{fig:introduction}(a), the majority of these approaches uniformly treat each layer and preserve the top-$B$ most important tokens at each respective layer.
However, it remains unclear whether the strategy of equally preserving the top-$B$ important tokens across all layers is an effective way to optimize information retention during the compression process.

To answer this question, we investigate the impact of token removal on information loss from the perspectives of attention mechanisms and hidden state outputs across different layers. Our empirical results reveal that the minimum required budget size varies across layers and models to maintain the same level of attention or hidden state output loss as in fullKV inference.

Building on these findings, we propose a novel KV caches compression method, called \zigzagkv, which minimizes information loss by dynamically allocating budget size for each layer. 
As illustrated in Figure~\ref{fig:introduction}(b), the key idea of \zigzagkv~ is to adjust the budget size based on the uncertainty of each layer. For instance, Layer 0, which exhibits more diffuse attention (i.e., higher uncertainty), is allocated a larger budget to retain its KV caches and reduce information loss. In contrast, layers with more concentrated attention (i.e., lower uncertainty) receive smaller budgets.
In practice, the proposed method first assigns an initial budget to each layer and then dynamically adjusts the remaining cache based on layer uncertainty. Experiments across various benchmarks demonstrate that \zigzagkv~ outperforms existing partialKV inference methods.

Our key contributions can be summarized as follows:
\begin{itemize}
    \item We analyze the differences in minimum budget sizes required to maintain information across layers and models, considering both attention mechanisms and hidden state outputs.

    \item Based on these observations, we propose \zigzagkv, a simple yet effective KV caches compression method that dynamically allocates budget size to each layer based on its uncertainty. 

    \item Experimental results show that \zigzagkv~ outperforms existing KV cache compression methods on two widely-used benchmarks: Needle-in-a-Haystack and LongBench.

\end{itemize}

\section{Problem Formulation}

\subsection{FullKV Inference}
Large Language Model (LLM) inference operates in an autoregressive manner. During training, the upper triangular part of the attention matrix is masked to ensure that each token only attends to itself and the preceding tokens. At inference time, a common approach is to cache the key-value vectors computed up to the current step and append the newly computed vectors to this cache.
Specifically, at each time step, the computed key states and value states are stored as a Key-Value (KV) Cache, which can be formalized as follows, where $h$ denotes the number of attention heads and $i \in [1, h]$ indexes these heads, $X$ represents the input embeddings, $W_i^K$ is the key projection matrix for head $i$, and $W_i^V$ is the value projection matrix for head $i$:
\begin{equation}
    K_i = XW_i^K,V_i=XW_i^V.
    \nonumber
\end{equation}
Then, for the computation of the next token, $x$ is mapped through the query projection $W_i^Q$, key projection $W_i^K$, and value projection $W_i^V$ as follows:
\begin{equation}
    q_i = xW_i^Q,k_i = xW_i^K,v_i = xW_i^V.
    \nonumber
\end{equation}
The key and value states are then updated based on the previous key-value (KV) Cache:
\begin{equation}
    K_i = Cat[K_i:k_i],V_i=Cat[V_i:v_i].
    \nonumber
\end{equation}
Finally, the updated query, key, and value states are used to compute the attention weights.
\begin{equation}
    A_i =  \text{Softmax}(\frac{q_{i}K_i^{T}}{\sqrt{d_h}}).
    \nonumber
\end{equation}
\begin{equation}
    y = \sum^{i \in [1, h]} A_i V_i W_i^O.
    \label{eq:fullkv}
\end{equation}
\subsection{PartialKV Inference}
In fullKV inference, the size of the key-value cache grows linearly with the total sequence length, which can lead to out-of-memory issues. Recent studies have shifted towards partialKV inference to address this.
Given a budget size of $B$ for each attention head, partialKV inference maintains the key-value cache by applying a cache eviction policy, as defined below:
\begin{equation}
    \hat{K_i} , \hat{V_i} = \text{Eviction}(K_i , V_i ).
    \nonumber
\end{equation}
Using the evicted key-value cache, attention weights are then calculated as follows:
\begin{equation}
    \hat{y} = \sum^{i \in [1, h]} \hat{A_i} \hat{V_i} W_i^O,
    \label{eq:partialkv}
\end{equation}
where $\hat{A_i} =  \text{Softmax}(\frac{q_{i}\hat{K_i}^{T}}{\sqrt{d_h}})$.

\section{Rethinking PartialKV Inference}
\label{partialkv_analysis}


Compared to fullKV inference, partialKV inference inevitably incurs some degree of information loss due to the reduction of the key-value (KV) cache. 
To mitigate this information loss, many partialKV inference strategies seek to minimize the discrepancy from fullKV inference with a fixed memory budget. For example, some approaches~\citep{zhang2023h,liu2024scissorhands,li2024snapkv} focus on retaining tokens with the highest attention scores, aiming to preserve the most crucial information from fullKV and thus reduce attention loss, which in turn helps minimize overall information loss. 
Typically, these methods heuristically assign the same budget size $B$ to each attention head across different layers, retaining the Top-$B$ most important tokens.

However, it is uncertain whether preserving the top-$B$ tokens with the highest scores equally across all heads in each layer effectively optimizes information retention.
In the following section, we will examine the impact of token removal on information loss, considering the attention mechanisms and hidden state outputs across different layers.

\paragraph{Layer-Specific Budget Setting for Attention Retention.}
\label{attention_loss}
\begin{figure}[t]
        \centering
	\includegraphics[width=0.9\linewidth]{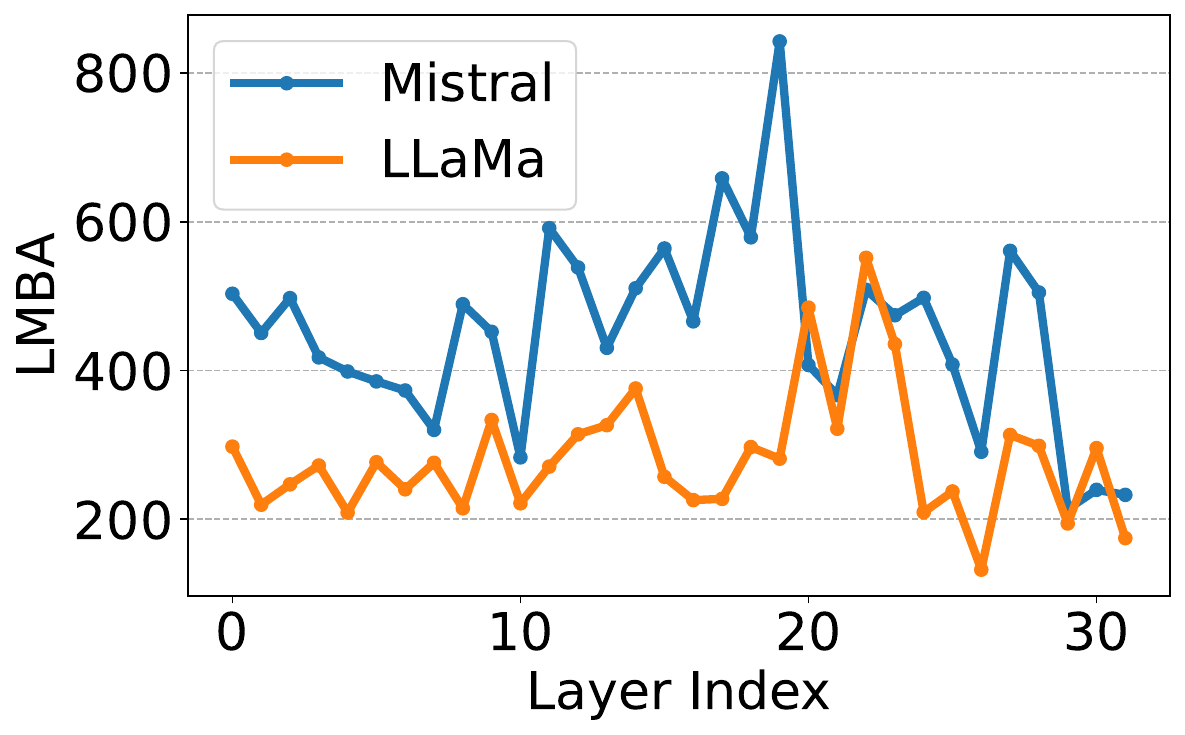}
	\caption{LMBA across various Layers of Mistral and LLaMa on 2WikiMQA dataset.} 
	\label{fig:lmba}
\end{figure}
Firstly, we investigate the relationship between the budget size to retain KV caches and information loss across different layers from the perspective of attention mechanisms. 
Specifically, we calculate the \textbf{M}inimum \textbf{B}udget size required to maintain 90\% of the total \textbf{A}ttention score (MBA) for each head, which corresponds to the attention loss as 0.1. This is formally defined as:
\begin{equation}
    \text{MBA} = \underset{I \subseteq [n]}{\arg \min} \, \left\{ |I| \, \big| \, 1.0 - \sum_{i \in I} a_i < 0.1 \right\}.
    \nonumber
\end{equation}
Next, we compute the average MBA across all heads within a layer to determine the required budget size for that layer, termed \textbf{L}ayer \textbf{M}inimum \textbf{B}udget size to maintain \textbf{A}ttention score (LMBA).
The LMBA is formally defined as:
\begin{equation}
    \text{LMBA} = \frac{1}{h} \sum_{i=1}^{h} \text{MBA}_i.
    \label{eq:lmba}
\end{equation}
A higher LMBA indicates that more tokens are required to maintain an attention loss of 0.1 in that layer, suggesting a larger budget allocation.
Empirically, we analyze the LMBA on two widely-used large language models, Mistral-7B-Instruct-v0.3~\citep{jiang2023mistral}(Mistral) and LLaMA-3.1-8B-Instruct~\citep{dubey2024llama}(LLaMA), using 200 samples from the 2WikiQA dataset~\citep{ho2020constructing}.
As illustrated in Figure~\ref{fig:lmba}, the LMBA varies across different layers: it initially requires a relatively larger budget in the lower layers to maintain an attention loss of 0.1, then decreases in the middle layers, increases again in the higher layers, and decreases.
This phenomenon indicates that the LMBA varies across different layers to maintain the same level of attention loss.
\textbf{{\em This suggest that applying a uniform budget size $B$ across all layers to retain the top-$B$ tokens may not be optimal for preserving attention information.}}



\paragraph{Layer-Specific Budget Setting for Hidden State Output Retention.}
\label{output_loss}
\begin{figure}[t]
        \centering
	\includegraphics[width=0.9\linewidth]{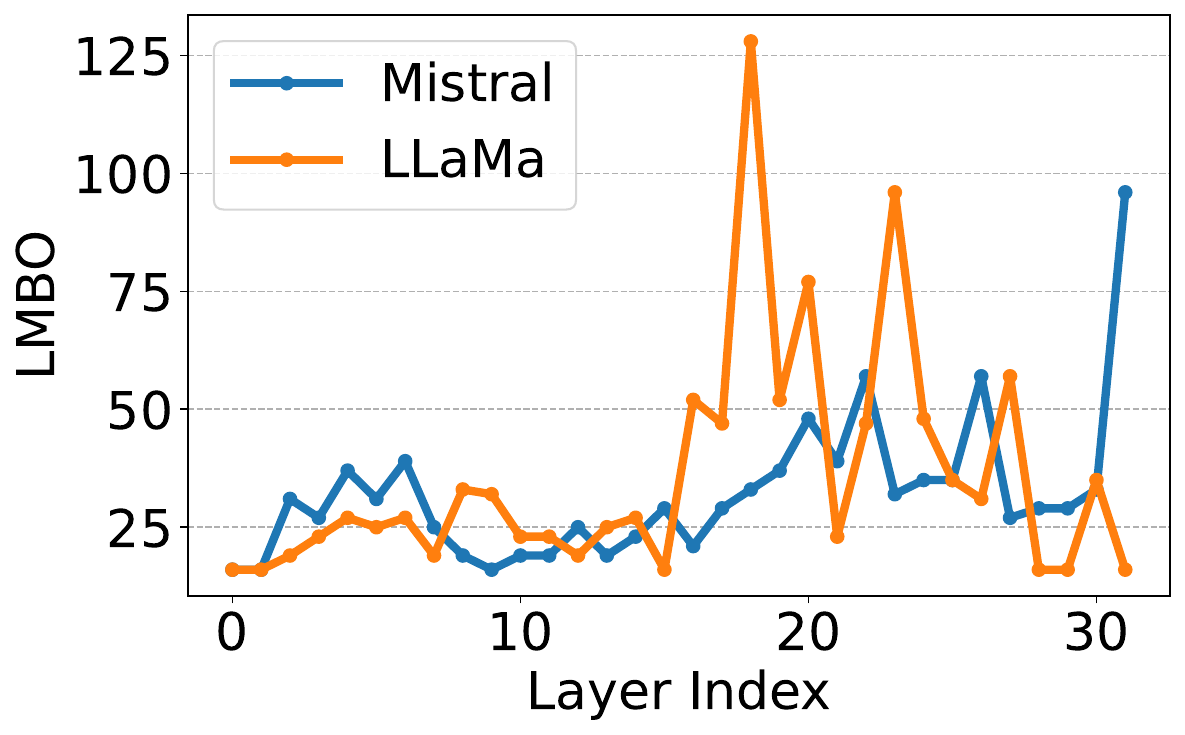}
	\caption{LMBO across various Layers of Mistral and LLaMa on 2WikiMQA dataset.} 
	\label{fig:lmbo}
\end{figure}
Next, we investigate the impact of budget size on information loss across different layers, focusing on the output of hidden states. 
Similarly, we examine each layer to determine the minimal budget size required for achieving a similarity of at least 90\% between partialKV and fullKV inference outputs. This threshold is denoted as the Layer-wise Minimum Budget for Output (LMBO). The formal definition of LMBO is as follows:
\begin{equation}
    \text{LMBO} = \arg \min_{B} \left\{ B \, \big| \, 1.0 - \text{similarity}(y, \hat{y}) < 0.1 \right\}.
    \nonumber
\end{equation}
Where $y$ represents the hidden state output in fullKV inference, as shown in Equation~\ref{eq:fullkv}, and $\hat{y}$ is the hidden state output inf partialKV inference, as illustrated in Equation~\ref{eq:partialkv}.
The experimental results, depicted in Figure \ref{fig:lmbo}, indicate that similar to LMBA, the LMBO varies across different layers.
In the case of LLaMa, the cache size required to maintain information stability is minimal initially but increases with layer depth.
Conversely, for Mistral, the trend of required cache size to preserve stability is consistently upward as layer depth increases.
This phenomenon suggests that LMBO varies across different layers and model types to maintain a consistent level of hidden state output information loss.
\textbf{{\em Therefore, it may not be optimal to apply a uniform budget size to retain the top-$B$ tokens across all layers for preserving hidden state output information.}}

\section{\zigzagkv}
\label{method}
\subsection{Dynamic Budget Allocation Based on Layer Uncertainty}
\label{dynamic}
The analysis in Section~\ref{partialkv_analysis} demonstrates that using a uniform budget size $B$ across all layers to select the top-$B$ tokens is suboptimal for retaining information, both in terms of attention and hidden state outputs. 
Most current partialKV methods use this uniform approach, which may lead to unnecessary information loss.
For instance, Figure 1(a) illustrates that certain layers, particularly the first one, may risk discarding important information. In contrast, layers where information is concentrated on specific tokens do not require the same budget size allocation.

To mitigate this issue, we introduce \zigzagkv, a dynamic method for allocating the budget size more effectively across layers to enhance information retention. Given an average budget size $B$, we determine the uncertainty in each layer $l$ using the Layer Minimum Budget size to maintain Attention (LMBA) as defined in Equation~\ref{eq:lmba}. 
The uncertainty is then used to adjust the budget size for each specific layer as described below:
\begin{equation} 
    \text{uncertainty}_l = \frac{\text{LMBA}_l}{\sum{i \in [1, L]} \text{LMBA}_i }.
    \label{uncertainty}
\end{equation} 
\begin{equation} 
    \hat{B}_l = B \cdot L \cdot \text{uncertainty}_l.
    \label{alloc1}
\end{equation}
Where $L$ represents the total number of layers. 
As illustrated in Figure 1(b), layers with higher uncertainty are allocated a larger portion of the budget, while those with less uncertainty receive a smaller share. 

Allocating the budget solely based on layer uncertainty can result in shallow budget sizes for layers with lower uncertainty, potentially leading to inadequate information retention. For example, if the LMBA value of one layer is significantly higher compared to others, Equation~\ref{alloc1} could allocate an excessively large budget to this layer, leaving the remaining layers with limited resources and potentially leading to information loss.
To resolve this, we propose a mechanism where a fixed minimum budget, $B_{\text{bound}}$, is allotted to each layer to protect against information degradation. 
The leftover budget is then distributed dynamically, informed by the layer uncertainty. The allocation strategy is formulated as follows:
\begin{equation}
    \hat{B}_l = B_{\text{bound}} + (B-B_{\text{bound}}) \cdot L \cdot \text{uncertainty}_l,
    \label{dynamic_budget}
\end{equation}
where $\text{uncertainty}_l$ is calculated as shown in Equation~\ref{uncertainty}. 
By incorporating $B_{\text{bound}}$, the method ensures that each layer receives a guaranteed minimum budget to preserve information while allowing dynamic adjustments to optimize information retention based on uncertainty.

\subsection{KV Cache Selection}
After determining the budget size for each layer, the next is to select the crucial tokens for each head of specific layers. The core concept of KV cache selection involves dynamically updating the KV cache by leveraging cumulative attention scores~\citep{zhang2023h,li2024snapkv,zhang2024pyramidkv,wan2024d2o}.
Following \citet{li2024snapkv} using cumulative attention scores of instruction tokens as the importance scores, we adopt a similar approach by using the cumulative attention scores of the last $w$ tokens to assign importance scores to the prefix tokens.
Specifically, given the budget size calculated by Equation~\ref{dynamic_budget}, for each attention head $h$, the importance score for retaining the $i$-th token in the KV cache, denoted as $s^h_i$, is computed as:
\begin{equation}
    s^h_i = \sum_{j\in [n-w,n]} A^h_{ij}
\end{equation}
Where $n$ represents the sequence length of the prompt, and $[n-w, n]$ represents the range of the last segment (instruction tokens) in the prompt.

\section{Experiments}
\subsection{Backbone LLMs} We compare the proposed method against several baselines using two open-source LLMs, specifically LLaMa-3.1-8B-Instruct~\citep{dubey2024llama} and Mistral-7B-Instruct-v0.3~\citep{jiang2023mistral}.
\subsection{Benchmarks}
The proposed approach is evaluated on two widely used benchmarks: Needle-in-a-Haystack and LongBench.
\paragraph{Needle-in-a-Haystack} The Needle-in-a-Haystack testing~\citep{needleinhaystack,fu2024data} challenges the model to accurately retrieve a specific sentence (the "needle") hidden within a lengthy document (the "haystack"), with the sentence placed at a random location. This test evaluates whether LLMs can extract key information from extensive texts and specifically examines the impact of the proposed adaptive allocation on the models’ long-context retrieval abilities. We evaluate all partial KV inference methods for this test using mean cache budgets $B \in \{128, 256, 512, 1024\}$.
\paragraph{LongBench} LongBench~\citep{bai2023longbench} is a multi-task benchmark designed to rigorously evaluate long-context understanding across various datasets, including single- and multi-document QA, summarization, few-shot learning, synthetic tasks, and code completion. For LongBench, we evaluate all partial KV inference methods using mean cache budgets $B \in \{128, 256, 512, 1024, 2048\}$.
\subsection{Baselines} We conduct experiments comparing the following methods:
\textbf{FullKV} (\fullkv) caches all keys and values for every input token in each layer.
\textbf{StreamingLLM} (\streamingllm)~\citep{xiao2023efficient} retains the KV cache of the last $\alpha$ tokens and the first $k-\alpha$ tokens.
\textbf{Heavy Hitter Oracle} (\ho)~\citep{zhang2023h} is a KV cache compression policy that dynamically balances recent and "Heavy Hitter" (H2) tokens. H2O maintains a fixed cache size across all layers.
\textbf{SnapKV} (\snapkv)~\citep{li2024snapkv} compresses KV caches by selecting and clustering important tokens for each attention head. Unlike H2O, SnapKV captures attention signals using patterns from a localized window and applies a more nuanced clustering algorithm, including a pooling layer.
\textbf{PyramidKV} (\pyramidkv)~\citep{zhang2024pyramidkv} proposes a layer-wise retention strategy that reduces cache size per layer based on depth. For KV cache selection, PyramidKV employs the same method as SnapKV.
\textbf{ZigZagKV} (proposed method) is detailed in Section \ref{method}.

\subsection{Main Results}

\begin{figure}[t]
\centering
\subfigure[Mistral-7B-Instruct-v0.3]{
\includegraphics[width=0.75\linewidth]{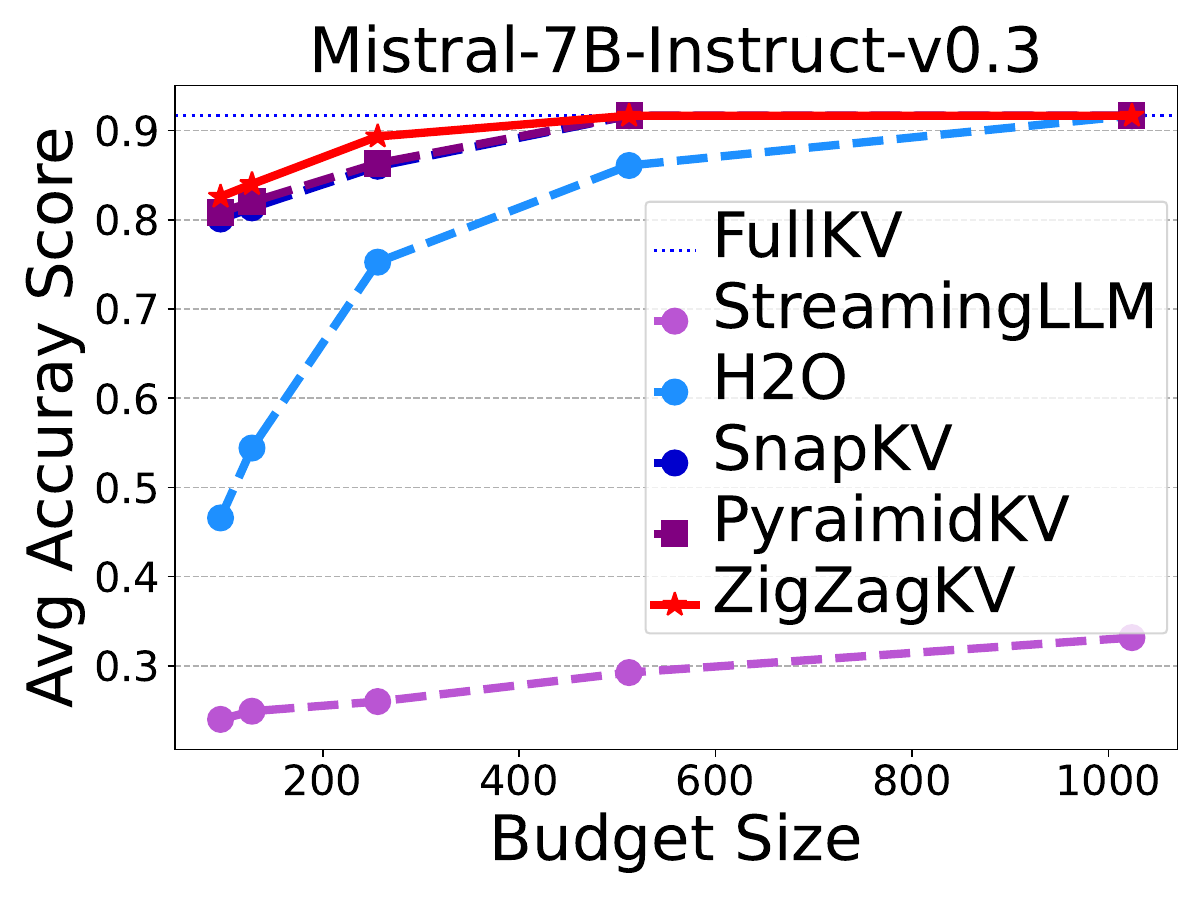}
}
\subfigure[LLaMa-3.1-8B-Instruct]{
\includegraphics[width=0.75\linewidth]{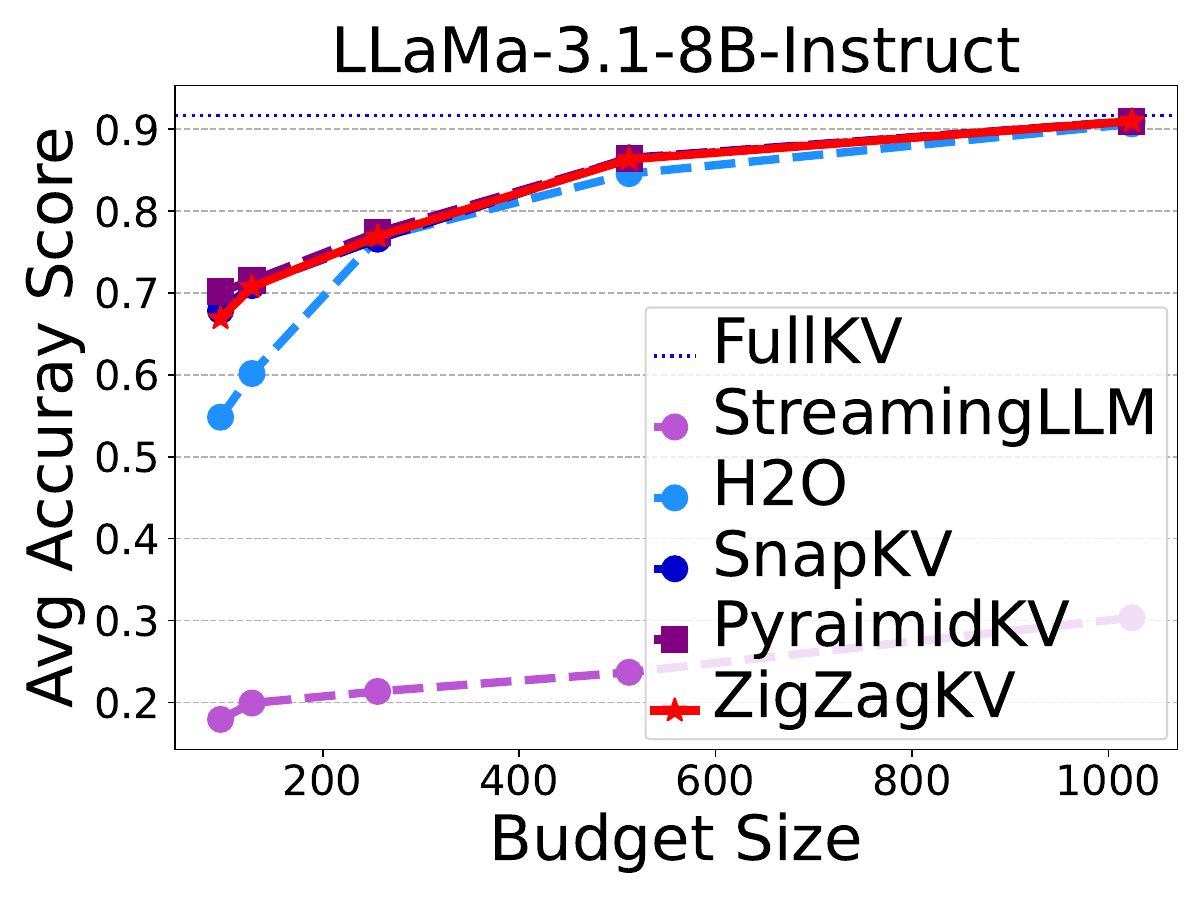}
}
\caption{The evaluation results from Needle-in-a-HayStack testing across 96, 128, 256, 512 and 1024 budget sizes on Mistral-7B-Instruct-v0.3 and LLaMa-3.1-8B-Instruct. Proposed method \zigzagkv \ outperforms \ho,\ \snapkv,\ \pyramidkv\ and \streamingllm, especially in limited budget sizes.}
\label{fig:needle_main}
\end{figure}

\begin{figure*}[t]

    \centering
    \subfigure[\streamingllm, Acc Score=26]{\includegraphics[width=0.3\textwidth]{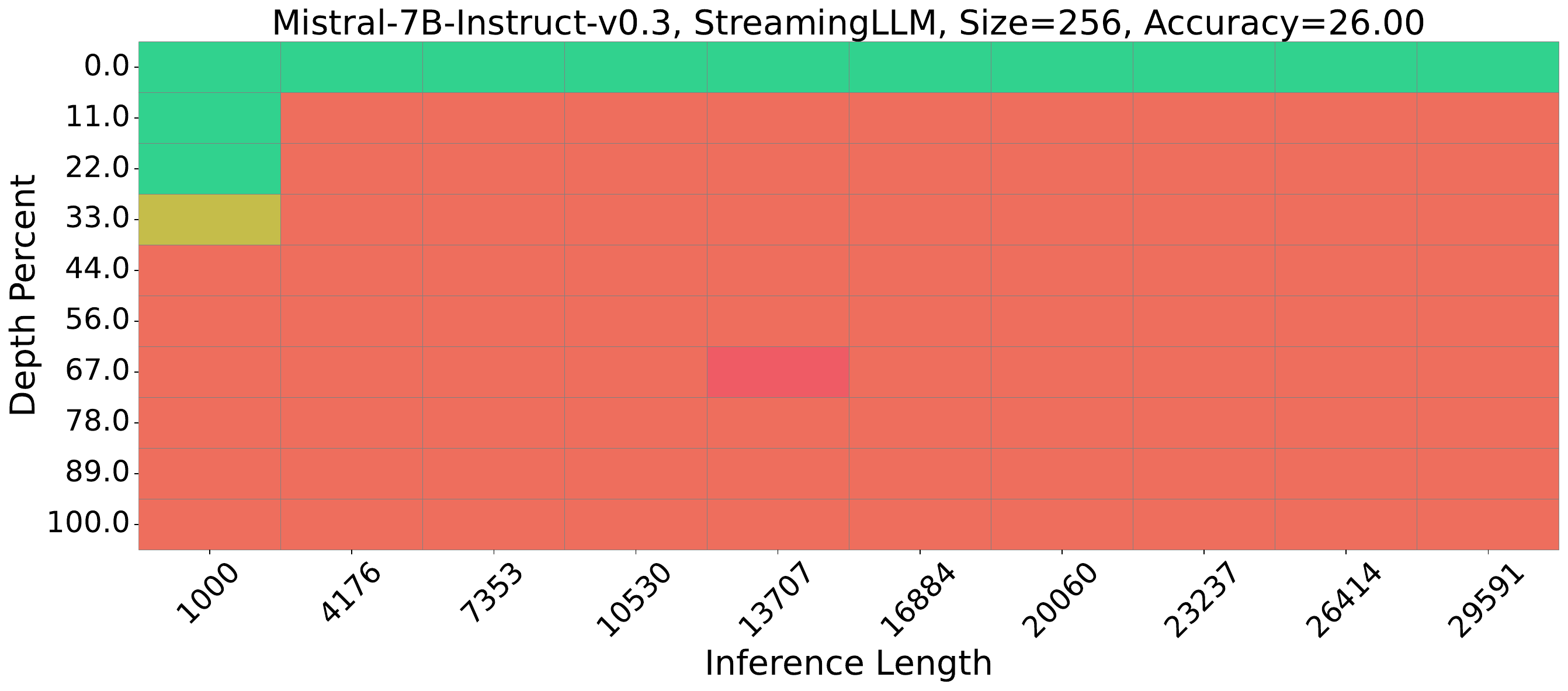}   }
    \subfigure[\ho, Acc Score=75.25]{\includegraphics[width=0.3\textwidth]{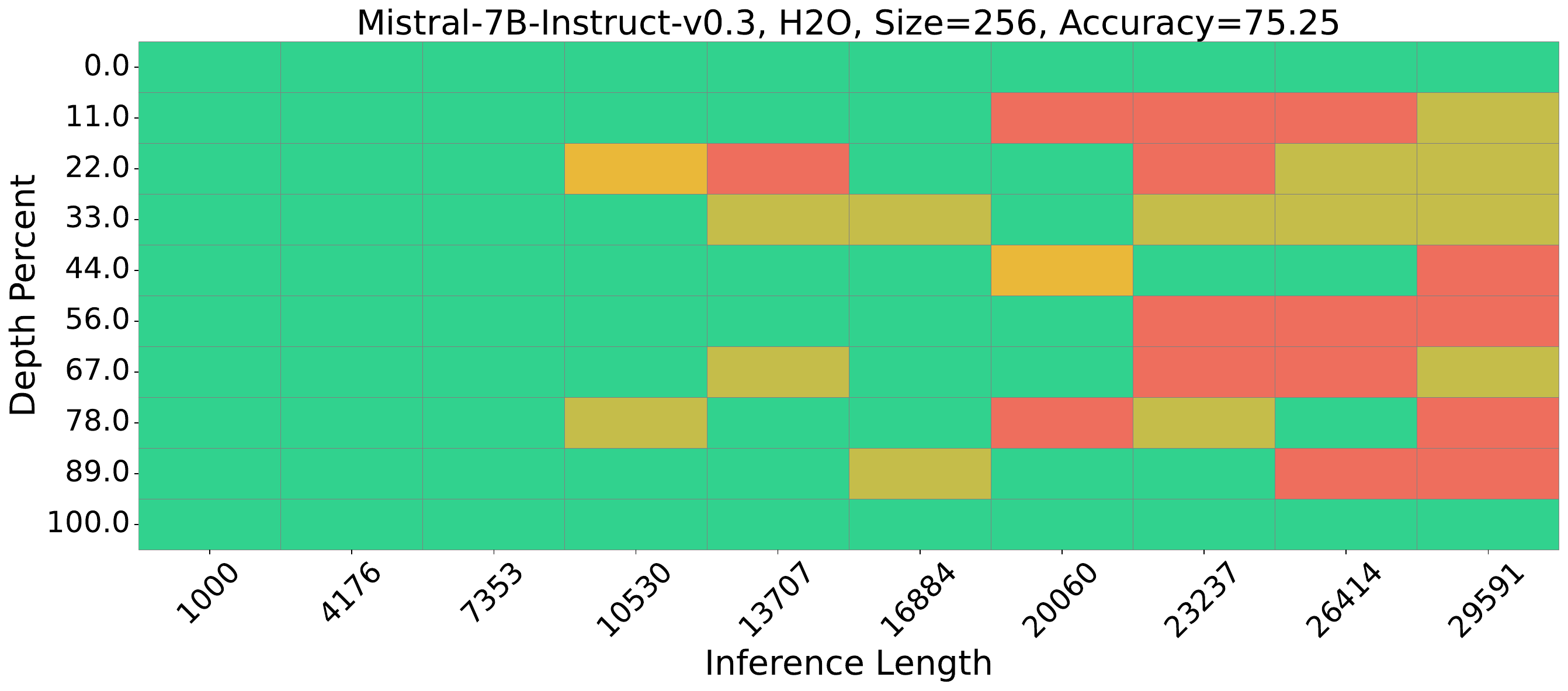}}
    \subfigure[\snapkv, Acc Score=86.00]{\includegraphics[width=0.3\textwidth]{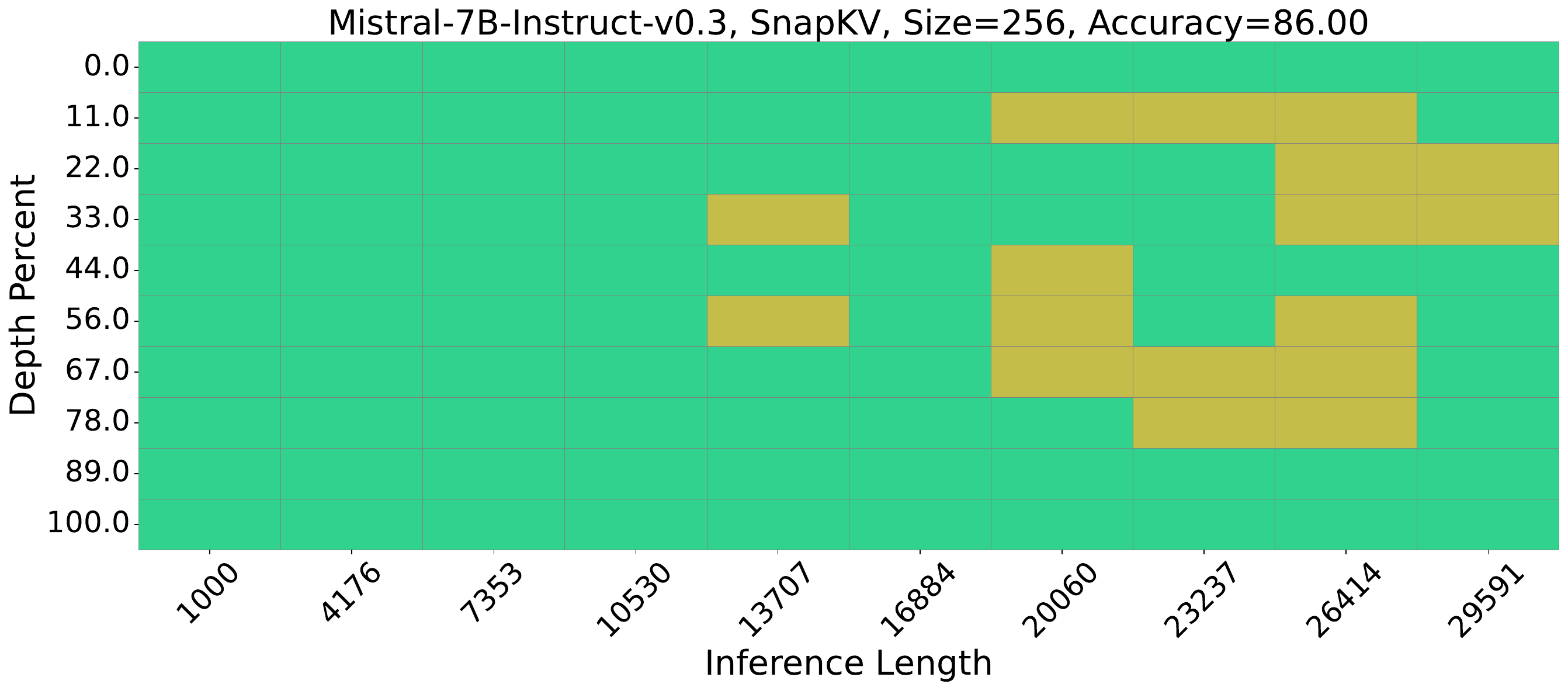}   }
    
    \subfigure[\pyramidkv, Acc Score=86.33]{\includegraphics[width=0.3\textwidth]{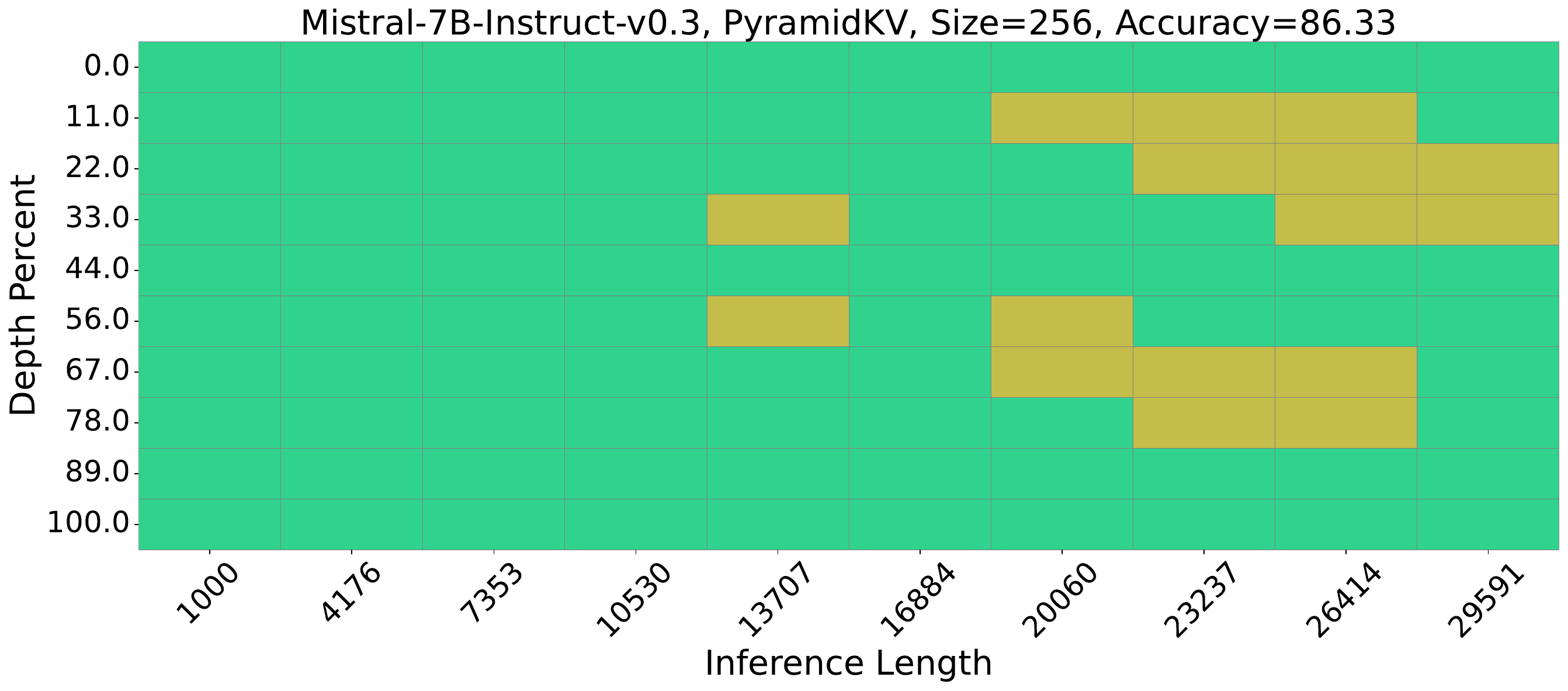}}
    \subfigure[\zigzagkv, Acc Score=89.33]{\includegraphics[width=0.3\textwidth]{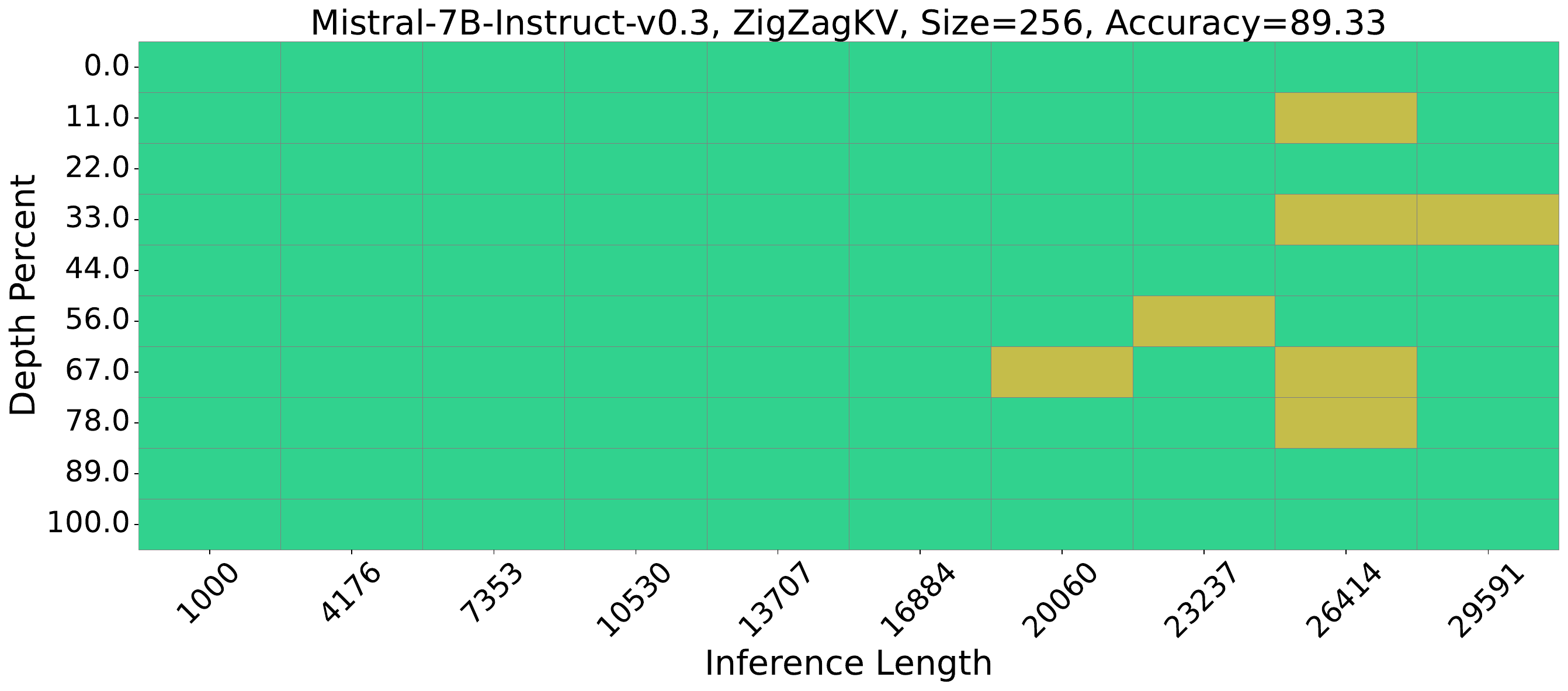}}
    \subfigure[\fullkv, Acc Score=91.67]{\includegraphics[width=0.3\textwidth]{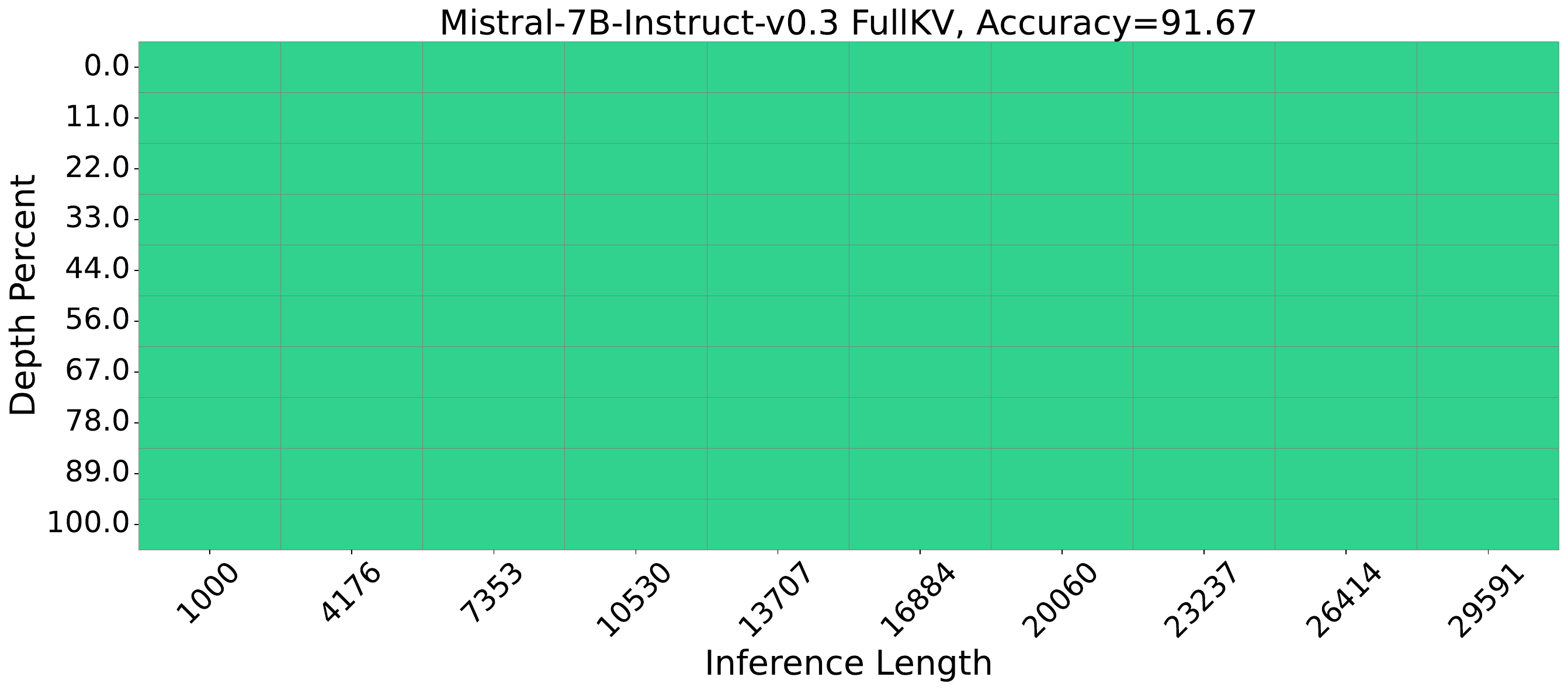}}

    \vspace{-3mm}

    \caption{Performance comparison for Needle-in-a-Haystack testing with a budget size of 256 on Mistral-7B-Instruct-v0.3 and LLaMa-3.1-8B-Instruct. The x-axis represents the length of the document, while the y-axis indicates the position where the needle is located. 
    A red cell indicates that the model fails to recall the information in the needle, whereas a green cell indicates success.}
    \label{fig:needle_256}

\end{figure*}

\begin{table*}[t!]
	\centering
	\addtolength{\tabcolsep}{-4.1pt}
		\renewcommand{\arraystretch}{0.7}
            \sisetup{detect-weight=true, round-mode=places, round-precision=1}
		\begin{tabular}{
				l@{}   c@{\hspace{-0.1ex}}c@{\hspace{0.2ex}}  c@{\hspace{0.2ex}}c@{\hspace{-2.2 ex}}c@{\hspace{-2ex}}c   @{\hspace{-0.5ex}}c@{\hspace{-1.2ex}}c@{\hspace{-1.2ex}}c@{\hspace{0.8ex}}   c@{\hspace{-0.2ex}}c@{\hspace{-1.2ex}} c@{\hspace{0.4ex}}c@{\hspace{0ex}}c@{\hspace{1.7ex}}c@{\hspace{0.3ex}}c@{\hspace{1.5ex}}c@{}
			}
   
			\toprule
			& \multicolumn{3}{c}{\small Single-Doc. QA}                                                                                                                                   & \multicolumn{3}{c}{\small Multi-Doc. QA}                                                                                                                                          & \multicolumn{3}{c}{\small Summarization}                                                                                                                                             & \multicolumn{3}{c}{\small Few-shotLearning}                                                                                                                                     & \multicolumn{2}{c}{\small Synthetic}                                                                                & \multicolumn{2}{c}{\small Code}                                                                                   & \multicolumn{1}{c}{}                                  \\ \cmidrule(lr){2-4}\cmidrule(lr){5-7}\cmidrule(lr){8-10}\cmidrule(lr){11-13}\cmidrule(lr){14-15}\cmidrule(lr){16-17}
			 & \small \rotatebox[origin=c]{-50}{NrtvQA} & \small \rotatebox[origin=c]{-50}{Qasper} & \small \rotatebox[origin=c]{-50}{MF-en} & \small \rotatebox[origin=c]{-50}{HotpotQA} & \small \rotatebox[origin=c]{-50}{2WikiMQA} & \small \rotatebox[origin=c]{-50}{Musique} & \small \rotatebox[origin=c]{-50}{GovReport} & \small \rotatebox[origin=c]{-50}{QMSum} & \small \rotatebox[origin=c]{-50}{MultiNews} & \small \rotatebox[origin=c]{-50}{TREC} & \small \rotatebox[origin=c]{-50}{TriviaQA} & \small \rotatebox[origin=c]{-50}{SAMSum} & \small \rotatebox[origin=c]{-50}{PCount} & \small \rotatebox[origin=c]{-50}{PRe} & \small \rotatebox[origin=c]{-50}{Lcc} & \multicolumn{1}{l}{\small \rotatebox[origin=c]{-50}{RB-P}} & \small \rotatebox[origin=c]{0}{\makecell{Avg.}}  \\ \midrule

            \small \fullkv       & \small\num{28.74} & \small\num{41.58} & \small\num{52.88} & \small\num{49.37} & \small\num{39.01} & \small\num{28.58} & \small\num{34.78} & \small\num{25.63} & \small\num{27.86} & \small \num{76} & \small\num{88.59} & \small\num{47.4} & \small\num{5.5} & \small \num{98} & \small\num{61.39} & \small\num{62.57}                & \multicolumn{1}{|l}{\small $\:$\num{47.99}}                    
   \\ \midrule
                        \multicolumn{18}{c}{\small   KV Size = 128} \\
   \small \streamingllm
   & \small\num{20.04} & \small\num{19.91} & \small\num{29.39} & \small\num{37.00} & \small\num{24.14} & \small\num{16.77} & \small\num{18.91} & \small\num{17.21} & \small\num{20.19} & \small\num{46.50} & \small\num{73.66} & \small\num{18.10} & \small\num{4.50} & \small\num{67.83} & \small\num{32.11}& \multicolumn{1}{l|} {{\small\num{36.60}}}& {\small$\:$\num{30.18}}                \\
   
                        \small \ho          & \small\num{25.33} & \small\num{29.12} & \small\num{44.56} & \small\num{45.67} & \small\num{32.88} & \small\num{23.80} & \small\num{21.91} & \small \textbf{\num{23.02}} & \small\num{21.17} & \small\num{34.00} & \small\num{88.19} & \small \textbf{\num{44.17}} & \small\num{6.00} & \small\num{94.00} & \small\num{53.73}& \multicolumn{1}{l|} {{\small\num{52.52}}}& {\small$\:$\num{40.00}}                \\

                        \small \snapkv       & \small\num{25.33} & \small\num{30.62} & \small \textbf{\num{48.23}} & \small\num{47.59} & \small\num{35.09} & \small\num{21.34} & \small\num{21.78} & \small\num{22.19} & \small\num{21.74} & \small\num{69.50} & \small\num{88.59} & \small\num{43.68} & \small\num{6.00} & \small\num{93.50} & \small\num{55.99}& \multicolumn{1}{l|} {{\small\textbf{\num{55.23}}}}& {\small$\:$\num{42.90}}                \\

  \small \pyramidkv      & \small\num{25.61} & \small \textbf{\num{31.48}} & \small\num{47.87} & \small \textbf{\num{47.94}} & \small\num{35.17} & \small \textbf{\num{25.36}} & \small\num{21.94} & \small\num{22.26} & \small\num{21.78} & \small \textbf{\num{70.00}} & \small\num{88.21} & \small\num{43.80} & \small\num{4.00} & \small \textbf{\num{94.50}} & \small\num{55.68}& \multicolumn{1}{l|} {{\small\num{54.98}}}& {\small$\:$\num{43.16}} \\

   \small \zigzagkv      & \small \textbf{\num{25.96}} & \small\num{30.43} & \small\num{48.06} & \small\num{47.71} & \small \textbf{\num{35.74}} & \small\num{24.70} & \small \textbf{\num{22.04}} & \small\num{22.55} & \small \textbf{\num{21.85}} & \small\num{69.50} & \small \textbf{\num{88.95}} & \small\num{44.04} & \small \textbf{\num{6.50}} & \small\num{93.50} & \small \textbf{\num{56.13}}& \multicolumn{1}{l|} {{\small\num{55.13}}}& {\small$\:$\textbf{\num{43.30}}} \\
   
   \midrule
                        \multicolumn{18}{c}{\small   KV Size =256}  \\
\small \streamingllm  & \small\num{20.06}  & \small\num{21.40}  & \small\num{32.52}  & \small\num{36.98}  & \small\num{25.43}  & \small\num{16.78}  & \small\num{21.84}  & \small\num{16.92}  & \small\num{23.58}  & \small\num{57.50}  & \small\num{72.38}  & \small\num{18.52}  & \small\num{4.00}  & \small\num{65.00}  & \small\num{34.36}& \multicolumn{1}{l|} {{\small\num{37.10}}}& {\small$\:$\num{31.52}} 
\\
\small \ho  & \small\num{24.67}  & \small\num{29.88}  & \small\num{47.01}  & \small\num{45.96}  & \small\num{35.46}  & \small\num{25.16}  & \small\num{22.91}  & \small \textbf{\num{24.08} }  & \small\num{22.75}  & \small\num{35.00}  & \small\num{88.56}  & \small\num{44.55}  & \small\num{4.50}  & \small \textbf{\num{96.50} }  & \small\num{56.43}& \multicolumn{1}{l|} {{\small\num{54.47}}}& {\small$\:$\num{41.12}} 
\\
\small \snapkv  & \small\num{26.76}  & \small\num{33.22}  & \small\num{52.17}  & \small\num{47.87}  & \small\num{37.50}  & \small\num{24.22}  & \small\num{23.80}  & \small\num{23.30}  & \small\num{23.36}  & \small\num{73.00}  & \small\num{88.79}  & \small\num{44.98}  & \small\num{5.00}  & \small\num{96.00}  & \small\num{58.29}& \multicolumn{1}{l|} {{\small\num{58.19}}}& {\small$\:$\num{44.78}} 
\\
\small \pyramidkv  & \small\num{26.71}  & \small \textbf{\num{33.99} }  & \small\num{51.61}  & \small\num{47.53}  & \small\num{38.05}  & \small \textbf{\num{27.36} }  & \small \textbf{\num{24.02} }  & \small\num{23.57}  & \small\num{23.48}  & \small \textbf{\num{74.00} }  & \small\num{88.67}  & \small \textbf{\num{44.99} }  & \small\num{5.00}  & \small\num{96.00}  & \small \textbf{\num{58.41} }& \multicolumn{1}{l|} {{\small\num{58.12}}}& {\small$\:$\num{45.09}} 
\\
\small \zigzagkv      & \small \textbf{\num{27.62} }  & \small\num{33.44}  & \small \textbf{\num{53.38} }  & \small \textbf{\num{48.63} }  & \small \textbf{\num{38.08} }  & \small\num{27.29}  & \small\num{23.99}  & \small\num{23.55}  & \small \textbf{\num{23.66} }  & \small\num{73.00}  & \small \textbf{\num{88.89} }  & \small\num{44.96}  & \small \textbf{\num{6.00} }  & \small\num{96.00}  & \small\num{58.25}& \multicolumn{1}{l|} {{\small\textbf{\num{58.48} }}}& {\small$\:$\textbf{\num{45.33} }} 
\\
   
   \midrule
                        \multicolumn{18}{c}{\small   KV Size = 512} \\

   \small \streamingllm  & \small\num{20.80}  & \small\num{22.16}  & \small\num{34.44}  & \small\num{37.86}  & \small\num{26.07}  & \small\num{16.08}  & \small\num{25.17}  & \small\num{18.46}  & \small \textbf{\num{26.48} }  & \small\num{65.50}  & \small\num{71.51}  & \small\num{18.11}  & \small\num{3.25}  & \small\num{67.50}  & \small\num{36.92}& \multicolumn{1}{l|} {{\small\num{37.51}}}& {\small$\:$\num{32.99}} 
\\
\small \ho  & \small\num{25.28}  & \small\num{33.38}  & \small\num{50.45}  & \small\num{48.37}  & \small \textbf{\num{39.26} }  & \small \textbf{\num{27.24} }  & \small\num{24.30}  & \small\num{24.00}  & \small\num{24.41}  & \small\num{39.50}  & \small\num{88.67}  & \small\num{46.06}  & \small\num{5.50}  & \small \textbf{\num{97.00} }  & \small\num{58.96}& \multicolumn{1}{l|} {{\small\num{57.16}}}& {\small$\:$\num{43.10}} 
\\
\small \snapkv  & \small\num{27.38}  & \small\num{36.42}  & \small\num{53.97}  & \small\num{49.74}  & \small\num{38.72}  & \small\num{26.71}  & \small\num{25.84}  & \small\num{24.48}  & \small\num{25.23}  & \small \textbf{\num{77.78} }  & \small\num{89.28}  & \small\num{46.68}  & \small\num{5.00}  & \small\num{94.50}  & \small\num{60.23}& \multicolumn{1}{l|} {{\small\textbf{\num{60.93} }}}& {\small$\:$\num{46.43}} 
\\
\small \pyramidkv  & \small\num{26.65}  & \small\num{36.05}  & \small\num{53.67}  & \small \textbf{\num{50.08} }  & \small\num{38.41}  & \small\num{26.97}  & \small\num{25.81}  & \small\num{24.52}  & \small\num{25.31}  & \small\num{74.50}  & \small\num{89.44}  & \small\num{46.43}  & \small\num{5.50}  & \small\num{96.00}  & \small\num{60.15}& \multicolumn{1}{l|} {{\small\num{60.45}}}& {\small$\:$\num{46.25}} 
\\
\small \zigzagkv      & \small \textbf{\num{27.78}}  & \small \textbf{\num{36.93} }  & \small \textbf{\num{54.21} }  & \small\num{49.74}  & \small\num{39.06}  & \small\num{26.88}  & \small \textbf{\num{25.89} }  & \small \textbf{\num{24.85} }  & \small\num{25.19}  & \small\num{75.00}  & \small \textbf{\num{89.44} }  & \small \textbf{\num{46.70} }  & \small \textbf{\num{5.50} }  & \small\num{96.50}  & \small \textbf{\num{60.34} }& \multicolumn{1}{l|} {{\small\num{60.69}}}& {\small$\:$\textbf{\num{46.54} }} 
\\
   
   \midrule
                        \multicolumn{18}{c}{\small   KV Size = 1024} \\

   \small \streamingllm  & \small\num{21.96}  & \small\num{28.10}  & \small\num{41.21}  & \small\num{37.93}  & \small\num{26.95}  & \small\num{17.16}  & \small\num{27.90}  & \small\num{19.89}  & \small \textbf{\num{27.21} }  & \small\num{71.50}  & \small\num{70.34}  & \small\num{19.02}  & \small\num{5.37}  & \small\num{69.00}  & \small\num{40.98}& \multicolumn{1}{l|} {{\small\num{38.29}}}& {\small$\:$\num{35.18}} 
\\
\small \ho  & \small \textbf{\num{27.28} }  & \small\num{34.80}  & \small\num{51.22}  & \small\num{49.19}  & \small\num{37.03}  & \small\num{26.75}  & \small\num{26.07}  & \small\num{24.94}  & \small\num{26.28}  & \small\num{48.00}  & \small \textbf{\num{89.27} }  & \small\num{46.45}  & \small\num{5.00}  & \small\num{97.50}  & \small\num{60.50}& \multicolumn{1}{l|} {{\small\num{58.58}}}& {\small$\:$\num{44.30}} 
\\
\small \snapkv  & \small\num{26.79}  & \small\num{37.78}  & \small\num{52.71}  & \small\num{49.23}  & \small\num{38.86}  & \small\num{28.06}  & \small\num{28.24}  & \small \textbf{\num{25.26} }  & \small\num{26.75}  & \small\num{76.00}  & \small\num{88.99}  & \small\num{46.20}  & \small\num{5.50}  & \small\num{97.50}  & \small\num{61.30}& \multicolumn{1}{l|} {{\small\num{62.15}}}& {\small$\:$\num{46.96}} 
\\
\small \pyramidkv  & \small\num{26.69}  & \small\num{37.68}  & \small\num{52.69}  & \small\num{49.33}  & \small \textbf{\num{38.88} }  & \small\num{27.94}  & \small\num{28.19}  & \small\num{24.97}  & \small\num{26.84}  & \small\num{76.00}  & \small\num{89.24}  & \small\num{46.40}  & \small\num{5.50}  & \small\num{97.50}  & \small\num{61.43}& \multicolumn{1}{l|} {{\small\num{61.89}}}& {\small$\:$\num{46.95}} 
\\
\small \zigzagkv  & \small\num{26.91}  & \small \textbf{\num{37.84} }  & \small \textbf{\num{53.35} }  & \small \textbf{\num{49.57} }  & \small\num{38.86}  & \small \textbf{\num{28.11} }  & \small \textbf{\num{28.57} }  & \small\num{25.10}  & \small\num{26.89}  & \small \textbf{\num{76.00} }  & \small\num{89.24}  & \small \textbf{\num{46.49} }  & \small \textbf{\num{5.50} }  & \small \textbf{\num{98.00} }  & \small \textbf{\num{61.53} }& \multicolumn{1}{l|} {{\small\textbf{\num{62.24} }}}& {\small$\:$\textbf{\num{47.14} }} 
\\

\midrule
                        \multicolumn{18}{c}{\small   KV Size = 2048} \\

\small \streamingllm  & \small\num{23.26}  & \small\num{31.84}  & \small\num{47.11}  & \small\num{37.98}  & \small\num{29.47}  & \small\num{18.91}  & \small\num{30.12}  & \small\num{20.18}  & \small\num{27.28}  & \small\num{73.00}  & \small\num{70.04}  & \small\num{19.04}  & \small\num{5.25}  & \small\num{72.17}  & \small\num{45.61}& \multicolumn{1}{l|} {{\small\num{39.77}}}& {\small$\:$\num{36.94}} 
\\
\small \ho  & \small\num{27.69}  & \small\num{38.81}  & \small\num{52.69}  & \small\num{49.29}  & \small\num{38.39}  & \small\num{27.40}  & \small\num{29.11}  & \small\num{25.07}  & \small\num{27.33}  & \small\num{63.50}  & \small\num{89.11}  & \small\num{46.97}  & \small\num{5.50}  & \small\num{98.00}  & \small\num{61.32}& \multicolumn{1}{l|} {{\small\num{61.11}}}& {\small$\:$\num{46.33}} 
\\
\small \snapkv  & \small\num{28.19}  & \small\num{40.51}  & \small \textbf{\num{53.06} }  & \small\num{49.72}  & \small\num{38.55}  & \small\num{28.29}  & \small\num{30.78}  & \small \textbf{\num{25.56} }  & \small\num{27.48}  & \small\num{75.50}  & \small\num{88.86}  & \small \textbf{\num{47.30} }  & \small\num{5.50}  & \small\num{98.00}  & \small \textbf{\num{62.02} }& \multicolumn{1}{l|} {{\small\num{61.99}}}& {\small$\:$\num{47.58}} 
\\
\small \pyramidkv  & \small \textbf{\num{28.22} }  & \small \textbf{\num{40.84} }  & \small\num{52.76}  & \small\num{49.72}  & \small \textbf{\num{38.75} }  & \small \textbf{\num{28.51} }  & \small\num{30.59}  & \small\num{25.51}  & \small\num{27.47}  & \small\num{75.50}  & \small\num{89.11}  & \small\num{47.23}  & \small\num{5.50}  & \small\num{98.00}  & \small\num{61.86}& \multicolumn{1}{l|} {{\small\num{62.18}}}& {\small$\:$\num{47.61}} 
\\
\small \zigzagkv  & \small\num{28.18}  & \small\num{40.75}  & \small\num{52.95}  & \small \textbf{\num{49.93} }  & \small\num{38.64}  & \small\num{28.47}  & \small \textbf{\num{30.84} }  & \small\num{25.38}  & \small \textbf{\num{27.64} }  & \small \textbf{\num{75.50} }  & \small \textbf{\num{89.11} }  & \small\num{47.20}  & \small \textbf{\num{5.50} }  & \small \textbf{\num{98.00} }  & \small\num{62.00}& \multicolumn{1}{l|} {{\small\textbf{\num{62.34} }}}& {\small$\:$\textbf{\num{47.65} }} 
\\

            \bottomrule
		\end{tabular}
		\caption{Comparison Based on Mistral-7B-Instruct-v0.3 Among 16 Datasets}
	\label{tab:longbench_mistral}
\end{table*}

\begin{table*}[t]
	\centering
	\addtolength{\tabcolsep}{-4.1pt}
		\renewcommand{\arraystretch}{0.7}
  \sisetup{detect-weight=true, round-mode=places, round-precision=1}
		\begin{tabular}{
				l@{}   c@{\hspace{-0.1ex}}c@{\hspace{0.2ex}}  c@{\hspace{0.2ex}}c@{\hspace{-2.2 ex}}c@{\hspace{-2ex}}c   @{\hspace{-0.5ex}}c@{\hspace{-1.2ex}}c@{\hspace{-1.2ex}}c@{\hspace{0.8ex}}   c@{\hspace{-0.2ex}}c@{\hspace{-1.2ex}} c@{\hspace{0.4ex}}c@{\hspace{0ex}}c@{\hspace{1.7ex}}c@{\hspace{0.3ex}}c@{\hspace{1.5ex}}c@{}
			}
			\toprule
			& \multicolumn{3}{c}{\small Single-Doc. QA}                                                                                                                                   & \multicolumn{3}{c}{\small Multi-Doc. QA}                                                                                                                                          & \multicolumn{3}{c}{\small Summarization}                                                                                                                                             & \multicolumn{3}{c}{\small Few-shotLearning}                                                                                                                                     & \multicolumn{2}{c}{\small Synthetic}                                                                                & \multicolumn{2}{c}{\small Code}                                                                                   & \multicolumn{1}{c}{}                                  \\ \cmidrule(lr){2-4}\cmidrule(lr){5-7}\cmidrule(lr){8-10}\cmidrule(lr){11-13}\cmidrule(lr){14-15}\cmidrule(lr){16-17}
			& \small \rotatebox[origin=c]{-50}{NrtvQA} & \small \rotatebox[origin=c]{-50}{Qasper} & \small \rotatebox[origin=c]{-50}{MF-en} & \small \rotatebox[origin=c]{-50}{HotpotQA} & \small \rotatebox[origin=c]{-50}{2WikiMQA} & \small \rotatebox[origin=c]{-50}{Musique} & \small \rotatebox[origin=c]{-50}{GovReport} & \small \rotatebox[origin=c]{-50}{QMSum} & \small \rotatebox[origin=c]{-50}{MultiNews} & \small \rotatebox[origin=c]{-50}{TREC} & \small \rotatebox[origin=c]{-50}{TriviaQA} & \small \rotatebox[origin=c]{-50}{SAMSum} & \small \rotatebox[origin=c]{-50}{PCount} & \small \rotatebox[origin=c]{-50}{PRe} & \small \rotatebox[origin=c]{-50}{Lcc} & \multicolumn{1}{l}{\small \rotatebox[origin=c]{-50}{RB-P}} & \small \rotatebox[origin=c]{0}{\makecell{Avg.}}  \\ \midrule

\small \fullkv       & \small \num{28.77} & \small \num{13.04} & \small \num{27.51} & \small \num{16.71} & \small \num{16.50} & \small \num{11.42} & \small \num{34.48} & \small \num{23.47} & \small \num{26.89} & \small \num{72.50} & \small \num{91.65} & \small \num{43.81} & \small \num{7.10} & \small \num{97.73} & \small \num{65.11}& \multicolumn{1}{l|} {{\small\num{58.74}}}& {\small$\:$\num{39.71}}                    
   \\ \midrule
                        \multicolumn{18}{c}{\small   KV Size = 128} \\
   \small \streamingllm  & \small \num{11.74}  & \small \num{4.94}  & \small \num{14.52}  & \small \num{9.95}  & \small \num{10.42}  & \small \num{5.91}  & \small \num{21.53}  & \small \num{17.91}  & \small \num{20.63}  & \small \num{43.50}  & \small \num{71.07}  & \small \num{17.28}  & \small \textbf{ \num{9.48} }  & \small \num{72.73}  & \small \num{40.50}& \multicolumn{1}{l|} {{\small\num{36.99}}}& {\small$\:$\num{25.57}} 
\\
\small \ho  & \small \textbf{ \num{23.06} }  & \small \num{7.68}  & \small \num{19.01}  & \small \num{13.20}  & \small \num{13.38}  & \small \num{8.06}  & \small \num{22.32}  & \small \num{22.14}  & \small \num{20.05}  & \small \num{39.00}  & \small \num{90.24}  & \small \textbf{ \num{40.73} }  & \small \num{7.46}  & \small \num{93.73}  & \small \num{56.61}& \multicolumn{1}{l|} {{\small\num{48.88}}}& {\small$\:$\num{32.85}} 
\\
\small \snapkv  & \small \num{21.25}  & \small \num{8.38}  & \small \num{20.60}  & \small \num{14.62}  & \small \num{14.11}  & \small \num{8.43}  & \small \num{22.18}  & \small \textbf{ \num{22.57} }  & \small \num{21.53}  & \small \num{62.00}  & \small \num{90.17}  & \small \num{39.95}  & \small \num{8.07}  & \small \num{92.79}  & \small \num{60.68}& \multicolumn{1}{l|} {{\small\num{50.58}}}& {\small$\:$\num{34.87}} 
\\
\small \pyramidkv  & \small \num{21.29}  & \small \textbf{ \num{8.72} }  & \small \num{21.03}  & \small \num{13.92}  & \small \num{13.75}  & \small \textbf{ \num{9.15} }  & \small \num{22.17}  & \small \num{22.53}  & \small \num{21.83}  & \small \textbf{ \num{63.50} }  & \small \num{90.30}  & \small \num{40.12}  & \small \num{7.96}  & \small \num{93.91}  & \small \num{58.69}& \multicolumn{1}{l|} {{\small\num{49.95}}}& {\small$\:$\num{34.93}} 
\\
\small \zigzagkv  & \small \num{21.37}  & \small \num{8.55}  & \small \textbf{ \num{21.55} }  & \small \textbf{ \num{14.95} }  & \small \textbf{ \num{14.43} }  & \small \num{8.71}  & \small \textbf{ \num{22.61} }  & \small \num{22.46}  & \small \textbf{ \num{22.09} }  & \small \num{62}  & \small \textbf{ \num{90.82} }  & \small \num{40.53}  & \small \num{8.67}  & \small \textbf{ \num{94.3} }  & \small \textbf{ \num{61.07} }& \multicolumn{1}{l|} {{\small\textbf{ \num{51.65} }}}& {\small$\:$\textbf{ \num{35.36} }} 
\\
   
   \midrule
                        \multicolumn{18}{c}{\small   KV Size =256}  \\
\small \streamingllm  & \small \num{14.00}  & \small \num{5.57}  & \small \num{14.84}  & \small \num{10.23}  & \small \num{9.65}  & \small \num{6.18}  & \small \num{23.82}  & \small \num{17.95}  & \small \num{22.90}  & \small \num{54.50}  & \small \num{70.61}  & \small \num{17.70}  & \small \textbf{ \num{8.85} }  & \small \num{76.70}  & \small \num{42.20}& \multicolumn{1}{l|} {{\small\num{37.41}}}& {\small$\:$\num{27.07}} 
\\
\small \ho  & \small \num{24.98}  & \small \num{7.79}  & \small \num{20.22}  & \small \num{14.66}  & \small \num{13.35}  & \small \num{9.05}  & \small \num{23.34}  & \small \num{22.75}  & \small \num{21.12}  & \small \num{39.00}  & \small \num{90.62}  & \small \num{41.52}  & \small \num{7.12}  & \small \num{93.03}  & \small \num{60.00}& \multicolumn{1}{l|} {{\small\num{49.77}}}& {\small$\:$\num{33.65}} 
\\
\small \snapkv  & \small \num{24.18}  & \small \num{9.36}  & \small \num{23.15}  & \small \num{15.06}  & \small \num{14.66}  & \small \num{9.19}  & \small \num{24.11}  & \small \num{23.05}  & \small \num{23.19}  & \small \num{70.00}  & \small \num{91.44}  & \small \num{41.05}  & \small \num{7.18}  & \small \textbf{ \num{95.94} }  & \small \num{61.99}& \multicolumn{1}{l|} {{\small\num{53.71}}}& {\small$\:$\num{36.70}} 
\\
\small \pyramidkv  & \small \num{24.37}  & \small \num{9.15}  & \small \textbf{ \num{23.35} }  & \small \num{14.71}  & \small \num{14.81}  & \small \num{9.29}  & \small \textbf{ \num{24.27} }  & \small \num{23.15}  & \small \num{23.28}  & \small \num{70.00}  & \small \num{91.44}  & \small \textbf{ \num{41.66} }  & \small \num{7.14}  & \small \num{95.75}  & \small \num{61.63}& \multicolumn{1}{l|} {{\small\num{53.07}}}& {\small$\:$\num{36.69}} 
\\
\small \zigzagkv  & \small \textbf{ \num{25.49} }  & \small \textbf{ \num{9.48} }  & \small \num{23.31}  & \small \textbf{ \num{15.19} }  & \small \textbf{ \num{14.9} }  & \small \textbf{ \num{9.86} }  & \small \num{24.19}  & \small \textbf{ \num{23.2} }  & \small \textbf{ \num{23.5} }  & \small \textbf{ \num{70} }  & \small \textbf{ \num{91.64} }  & \small \num{41.61}  & \small \num{7.49}  & \small \num{94.5}  & \small \textbf{ \num{62.1} }& \multicolumn{1}{l|} {{\small\textbf{ \num{53.83} }}}& {\small$\:$\textbf{ \num{36.89} }} 
\\
   
   \midrule
                        \multicolumn{18}{c}{\small   KV Size = 512} \\

\small \streamingllm  & \small \num{12.82}  & \small \num{6.36}  & \small \num{19.35}  & \small \num{10.63}  & \small \num{10.05}  & \small \num{6.28}  & \small \num{25.87}  & \small \num{18.95}  & \small \num{24.72}  & \small \num{60.50}  & \small \num{71.94}  & \small \num{18.69}  & \small \textbf{ \num{8.06} }  & \small \num{79.62}  & \small \num{43.94}& \multicolumn{1}{l|} {{\small\num{38.85}}}& {\small$\:$\num{28.54}} 
\\
\small \ho  & \small \num{23.88}  & \small \num{8.54}  & \small \num{21.46}  & \small \num{14.29}  & \small \num{13.64}  & \small \num{9.57}  & \small \num{24.33}  & \small \num{22.61}  & \small \num{23.36}  & \small \num{41.00}  & \small \num{91.63}  & \small \num{41.46}  & \small \num{7.62}  & \small \num{94.32}  & \small \num{61.48}& \multicolumn{1}{l|} {{\small\num{51.67}}}& {\small$\:$\num{34.43}} 
\\
\small \snapkv  & \small \num{25.18}  & \small \textbf{ \num{11.26} }  & \small \num{25.10}  & \small \num{15.09}  & \small \textbf{ \num{15.65} }  & \small \textbf{ \num{9.89} }  & \small \num{26.05}  & \small \num{23.11}  & \small \textbf{ \num{24.72} }  & \small \num{70.50}  & \small \num{91.73}  & \small \num{41.42}  & \small \num{7.72}  & \small \num{96.17}  & \small \num{63.80}& \multicolumn{1}{l|} {{\small\textbf{ \num{55.58} }}}& {\small$\:$\num{37.69}} 
\\
\small \pyramidkv  & \small \num{25.94}  & \small \num{11.08}  & \small \num{24.70}  & \small \textbf{ \num{15.54} }  & \small \num{15.46}  & \small \num{9.78}  & \small \num{26.09}  & \small \num{23.29}  & \small \num{24.62}  & \small \num{70.50}  & \small \textbf{ \num{91.90} }  & \small \textbf{ \num{41.72} }  & \small \num{7.83}  & \small \num{96.33}  & \small \num{63.69}& \multicolumn{1}{l|} {{\small\num{54.87}}}& {\small$\:$\num{37.71}} 
\\
\small \zigzagkv  & \small \textbf{ \num{26.11} }  & \small \num{11.24}  & \small \textbf{ \num{25.19} }  & \small \num{15.48}  & \small \num{15.34}  & \small \num{9.61}  & \small \textbf{ \num{26.28} }  & \small \textbf{ \num{23.46} }  & \small \num{24.57}  & \small \textbf{ \num{70.5} }  & \small \num{91.73}  & \small \num{41.5}  & \small \num{8.05}  & \small \textbf{ \num{96.75} }  & \small \textbf{ \num{64.23} }& \multicolumn{1}{l|} {{\small\num{55.23}}}& {\small$\:$\textbf{ \num{37.83} }} 
\\
   
   \midrule
                        \multicolumn{18}{c}{\small   KV Size = 1024} \\

\small \streamingllm  & \small \num{12.97}  & \small \num{7.44}  & \small \num{20.93}  & \small \num{12.09}  & \small \num{10.64}  & \small \num{7.09}  & \small \num{27.79}  & \small \num{19.30}  & \small \textbf{ \num{25.96} }  & \small \num{67.50}  & \small \num{74.03}  & \small \num{19.49}  & \small \num{8.06}  & \small \num{79.91}  & \small \num{45.84}& \multicolumn{1}{l|} {{\small\num{39.04}}}& {\small$\:$\num{29.88}} 
\\
\small \ho  & \small \num{24.74}  & \small \num{9.95}  & \small \num{24.13}  & \small \num{14.84}  & \small \num{14.90}  & \small \num{9.86}  & \small \num{26.09}  & \small \num{23.24}  & \small \num{25.51}  & \small \num{45.00}  & \small \num{91.66}  & \small \num{42.36}  & \small \textbf{ \num{8.11} }  & \small \num{95.03}  & \small \num{63.32}& \multicolumn{1}{l|} {{\small\num{54.60}}}& {\small$\:$\num{35.83}} 
\\
\small \snapkv  & \small \textbf{ \num{28.84} }  & \small \num{11.78}  & \small \textbf{ \num{27.29} }  & \small \num{15.77}  & \small \num{15.60}  & \small \num{10.80}  & \small \num{28.32}  & \small \num{23.61}  & \small \num{25.77}  & \small \num{70.00}  & \small \num{91.73}  & \small \num{43.00}  & \small \num{7.41}  & \small \num{97.61}  & \small \num{63.92}& \multicolumn{1}{l|} {{\small\num{56.56}}}& {\small$\:$\num{38.63}} 
\\
\small \pyramidkv  & \small \num{28.23}  & \small \num{11.74}  & \small \num{26.85}  & \small \num{16.14}  & \small \num{15.70}  & \small \textbf{ \num{10.86} }  & \small \textbf{ \num{28.46} }  & \small \num{23.72}  & \small \num{25.82}  & \small \num{70.00}  & \small \num{91.73}  & \small \textbf{ \num{43.10} }  & \small \num{7.39}  & \small \num{97.78}  & \small \num{63.88}& \multicolumn{1}{l|} {{\small\num{56.86}}}& {\small$\:$\num{38.64}} 
\\
\small \zigzagkv  & \small \num{28.82}  & \small \textbf{ \num{12} }  & \small \num{26.81}  & \small \textbf{ \num{16.15} }  & \small \textbf{ \num{15.8} }  & \small \num{10.66}  & \small \num{28.38}  & \small \textbf{ \num{23.84} }  & \small \num{25.83}  & \small \textbf{ \num{70} }  & \small \textbf{ \num{91.73} }  & \small \num{43.07}  & \small \num{7.75}  & \small \textbf{ \num{97.82} }  & \small \textbf{ \num{63.95} }& \multicolumn{1}{l|} {{\small\textbf{ \num{56.86} }}}& {\small$\:$\textbf{ \num{38.72} }} 
\\

\midrule
                        \multicolumn{18}{c}{\small   KV Size = 2048} \\

\small \streamingllm  & \small \num{13.59}  & \small \num{10.07}  & \small \num{23.37}  & \small \num{11.66}  & \small \num{12.46}  & \small \num{7.24}  & \small \num{29.85}  & \small \num{19.75}  & \small \textbf{ \num{26.73} }  & \small \num{68.50}  & \small \num{79.50}  & \small \num{21.11}  & \small \textbf{ \num{8.21} }  & \small \num{64.25}  & \small \num{54.06}& \multicolumn{1}{l|} {{\small\num{40.38}}}& {\small$\:$\num{30.67}} 
\\
\small \ho  & \small \num{28.00}  & \small \num{11.32}  & \small \num{25.47}  & \small \num{16.04}  & \small \num{15.29}  & \small \num{10.35}  & \small \num{28.73}  & \small \num{23.28}  & \small \num{26.60}  & \small \num{56.50}  & \small \textbf{ \num{91.57} }  & \small \textbf{ \num{43.00} }  & \small \num{7.87}  & \small \num{96.73}  & \small \num{64.69}& \multicolumn{1}{l|} {{\small\num{57.22}}}& {\small$\:$\num{37.67}} 
\\
\small \snapkv  & \small \num{29.18}  & \small \num{12.44}  & \small \textbf{ \num{27.15} }  & \small \num{16.60}  & \small \num{16.26}  & \small \num{11.29}  & \small \num{30.43}  & \small \num{23.51}  & \small \num{26.58}  & \small \num{71.00}  & \small \num{91.48}  & \small \num{42.79}  & \small \num{7.70}  & \small \textbf{ \num{97.73} }  & \small \num{64.90}& \multicolumn{1}{l|} {{\small\num{58.22}}}& {\small$\:$\num{39.20}} 
\\
\small \pyramidkv  & \small \num{29.19}  & \small \num{12.42}  & \small \num{27.11}  & \small \num{16.58}  & \small \num{16.45}  & \small \textbf{ \num{11.59} }  & \small \num{30.60}  & \small \num{23.60}  & \small \num{26.32}  & \small \num{71.00}  & \small \num{91.48}  & \small \num{42.53}  & \small \num{7.70}  & \small \num{97.56}  & \small \num{64.73}& \multicolumn{1}{l|} {{\small\num{58.28}}}& {\small$\:$\num{39.20}} 
\\
\small \zigzagkv  & \small \textbf{ \num{29.38} }  & \small \textbf{ \num{12.73} }  & \small \num{27.13}  & \small \textbf{ \num{16.63} }  & \small \textbf{ \num{16.5} }  & \small \num{11.53}  & \small \textbf{ \num{30.79} }  & \small \textbf{ \num{23.71} }  & \small \num{26.65}  & \small \textbf{ \num{71} }  & \small \num{91.48}  & \small \num{42.66}  & \small \num{7.47}  & \small \num{97.62}  & \small \textbf{ \num{64.97} }& \multicolumn{1}{l|} {{\small\textbf{ \num{58.36} }}}& {\small$\:$\textbf{ \num{39.29} }} 
\\
   
            \bottomrule
		\end{tabular}
		\caption{Comparison Based on LLaMA-3.1-8B-Instruct Among 16 Datasets}
	\label{tab:longbench_llama}
\end{table*}

\begin{figure}[!htbp]
\centering
\subfigure[Mistral]{
\includegraphics[width=0.9\linewidth]{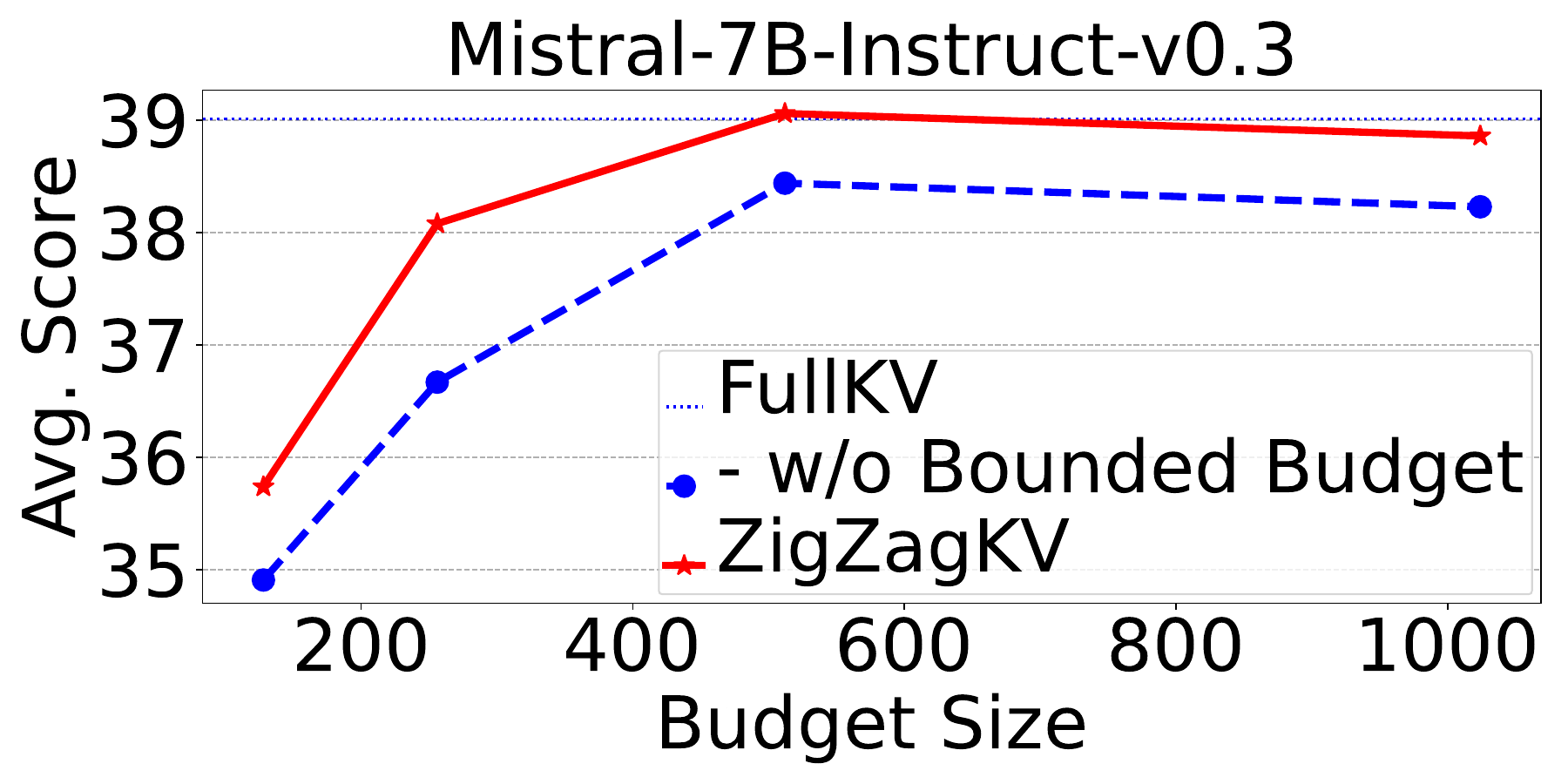}
}
\subfigure[LLaMa]{
\includegraphics[width=0.9\linewidth]{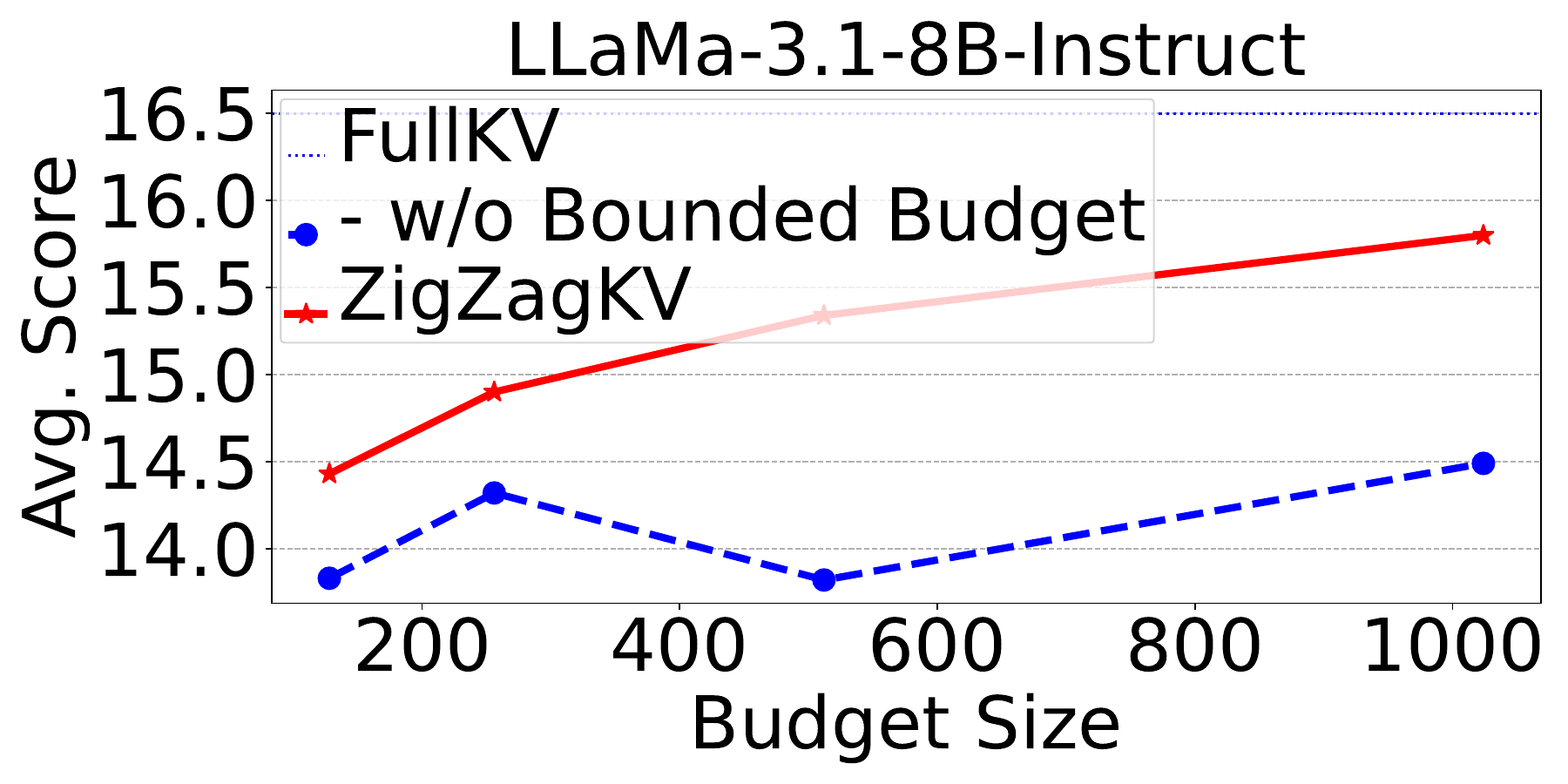}
}
\caption{Ablation studies between \zigzagkv~ w/ Bounded Budget and w/o Bounded Budget on both Mistral and LLaMa on 2WikiMQA Dataset.}
\label{fig:longbench_abs}
\end{figure}

\begin{table}[!htbp]
\centering
\sisetup{detect-weight=true, round-mode=places, round-precision=2}
\resizebox{1.0\linewidth}{!}{
\begin{tabular}{l|ccc|ccc}
\toprule
\multicolumn{1}{l}{Model} & \multicolumn{3}{c}{\bf Mistral} & \multicolumn{3}{c}{\bf LLaMa} \\ 
\midrule
Budget & 128 & 256 & 512 & 128 & 256 & 512       \\ 
\midrule
\snapkv & \num{2.706} & \num{1.543} & \num{0.893} & \num{1.763} & \num{0.901} & \num{0.463}  \\
\pyramidkv& \num{2.960} & \num{1.592} & \num{0.885} & \num{1.912} & \num{0.879} & \num{0.424} \\ 
\zigzagkv & \textbf{\num{2.447}} & \textbf{\num{1.249}} & \textbf{\num{0.611}} & \textbf{\num{1.504}} & \textbf{\num{0.637}} & \textbf{\num{0.226}}   \\
\bottomrule
\end{tabular}}
\caption{Attention loss of Mistral-7B-Instruct-v0.3 and LLaMa-3.1-8B-Instruct on 2WikiMQA Dataset.}
\label{table:attention_loss}
\end{table}

\begin{table}[!htbp]
\centering
\sisetup{detect-weight=true, round-mode=places, round-precision=2}
\resizebox{1.0\linewidth}{!}{
\begin{tabular}{l|ccc|ccc}
\toprule
\multicolumn{1}{l}{Model} & \multicolumn{3}{c}{\bf Mistral} & \multicolumn{3}{c}{\bf LLaMa} \\ 
\midrule
Budget & 128 & 256 & 512 & 128 & 256 & 512       \\ 
\midrule
\snapkv &  \num{2.550} & \num{1.509} & \num{0.851} & \textbf{\num{2.914}} & \num{1.673} & \num{0.884}  \\
\pyramidkv & \num{2.977} & \num{1.595} & \num{0.870} & \num{3.256} & \num{1.755} & \num{0.903} \\ 
\zigzagkv & \textbf{\num{2.544}} & \textbf{\num{1.495}} & \textbf{\num{0.830}} & \num{2.918} & \textbf{\num{1.668}} & \textbf{\num{0.878}} \\
\bottomrule
\end{tabular}}
\caption{Hidden state loss of Mistral-7B-Instruct-v0.3 and LLaMa-3.1-8B-Instruct on 2WikiMQA Dataset.}
\label{table:output_loss}
\end{table}

\paragraph{Results on Needle-in-a-Haystack Testing}
We first compare the proposed method with previous approaches on the Needle-in-a-Haystack test, as shown in Figure~\ref{fig:needle_main} and Figure~\ref{fig:needle_256}. \zigzagkv~ consistently outperforms previous methods under almost all constrained cache budget settings, particularly when the average budget is limited.
When the mean budget size is 256, \zigzagkv~ achieves an accuracy of 89.33\%, closely matching the retrieval accuracy of \fullkv~. In contrast, \streamingllm~ and \ho~ show a lower performance. Notably, \zigzagkv~ only requires an average budget of 256, even for 30K context, while \fullkv~ requires retaining the entire KV cache to inference.

\paragraph{Results on LongBench}
To assess the performance of the proposed method across various tasks, we conduct experiments using LongBench. The results are depicted in Table~\ref{tab:longbench_mistral} and Table~\ref{tab:longbench_llama}.
Similarly, \zigzagkv~ demonstrates improvements over the four baseline methods, achieving higher average scores across multiple tasks.
In particular, \zigzagkv~ outperforms \fullkv~ using only a mean KV cache size of 128 on the TriviaQA few-shot learning task. 
This demonstrates that the proposed method reduces memory overhead and captures more information from few-shot examples, highlighting its potential for further study in in-context learning tasks.

\subsection{Analysis and Ablation Studies}
\paragraph{\zigzagkv~Preserves More Attention Information.}
To investigate whether the proposed method achieves more attention information, as described in Section~\ref{attention_loss}, we calculated the average deviation from achieving 0.9 attention across various cache settings, which we refer to as attention loss. The results are presented in Table~\ref{table:attention_loss}.
Both \snapkv~and \pyramidkv~exhibit higher attention loss. This is because they either apply uniform treatment for all layers or allocate smaller cache budgets to higher layers, resulting in a significant deviation in attention scores compared to \fullkv.
In contrast, \zigzagkv~substantially reduces attention loss in the Mistral and LLaMa models, minimizing the attention score gap between the proposed method and \fullkv. This indicates that \zigzagkv~preserves more attention information compared to the baseline methods.

\paragraph{\zigzagkv~Maintains More Hidden State Information.}
Furthermore, to analyze whether the proposed method maintains a more stable hidden state output as described in Section~\ref{output_loss}, we compared the difference between the output of PartialKV Inference and FullKV Inference using the metric $1 - \text{similarity}(y, \hat{y})$, termed as output loss. We then computed the average output loss for each layer.
The results are illustrated in Table~\ref{table:output_loss}.
The mean output loss of \zigzagkv~is the lowest among all methods and models, except when the mean budget is set to 128.
This indicates that, compared to baseline methods, the proposed method effectively maintains more stable output by setting the budget size based on layer uncertainty.

\paragraph{Ablation Studies on Bounded Budget.}
To evaluate the effectiveness of the bounded budget operation in our proposed method, we compare the performance of our method with and without using this strategy.
As shown in Figure~\ref{fig:longbench_abs}, utilizing the bounded budget strategy enhances performance on both Mistral and LLaMa across various budget sizes.

\paragraph{Computational Overhead.}
\begin{table}[t]
    \centering
    \begin{tabular}{l c}
        \hline
        \textbf{Method} & \textbf{Average Latency (s)} \\
        \hline
        StreamingLM & 4.59 \\
        PyramidKV & 6.56 \\
        ZigZagKV & 6.50 \\
        \hline
    \end{tabular}
    \caption{Average Latency on NarrativeQA of LLaMA-3.1-8B-Instruct}
    \label{tab:latency}
\end{table}

To evaluate the computational cost differences between the \zigzagkv~ and the baseline method, we measure the latency of \streamingllm, \pyramidkv, and \zigzagkv, as presented in Table~\ref{tab:latency}. The latency tests indicate that \pyramidkv~ and \zigzagkv~ demonstrate similar performance. In contrast, \streamingllm~ exhibits faster processing speeds, while \streamingllm~ is faster but has a performance drop.

\section{Related Work}

Existing KV cache compression techniques can be broadly divided into two categories: fixed policies and adaptive policies.

For fixed policies, \citet{xiao2023efficient} and \citet{han2024lm} suggest that the initial tokens often receive consistently high attention weights across layers and heads. Therefore, they propose reducing the memory required for the KV cache by retaining only the first few tokens and local tokens. 
For adaptive policies
, most approaches select important tokens based on attention weights. \citet{liu2024scissorhands} introduce the "persistence of importance" hypothesis, suggesting that tokens with significant influence at one step will continue to impact future generations. \citet{zhang2023h,oren2024transformers} employ cumulative normalized attention scores to determine which tokens to retain while preserving recent tokens due to their strong correlation with the current generation. 
\citet{li2024snapkv} compress the KV cache by selecting and clustering necessary tokens based on the attention scores from the last segment of tokens.

While these methods differ in selecting tokens for KV cache retention, they generally apply a uniform budget size across layers, even though the optimal budget size may vary. 
Recently, some studies have explored budget size allocation across different layers~\citep{zhang2024pyramidkv,yang2024pyramidinfer,wan2024d2o}, but these approaches overlook the need for a minimum budget size to preserve essential information.

Unlike \pyramidkv, which heuristically allocates more cache in the lower layers and less in the higher ones, \zigzagkv~ leverages layer uncertainty to allocate the cache budget. As illustrated in Figure 2 of the submission, the largest  LMBA might not occur in the highest layers, which can lead to more cache being allocated to the middle layers.
In \pyramidkv, the cache sizes for all intermediate layers are set according to an arithmetic sequence. In contrast, with \zigzagkv~, the cache sizes for all layers vary depending on the context and models.

\section{Conclusion}
In this paper, we pay attention to the variation in minimum budget sizes required to retain information across different layers. A comprehensive analysis reveals that the necessary budget size differs across layers from the perspectives of attention mechanisms and hidden state outputs.
Building on these findings, we propose a training-free method that dynamically allocates budget sizes based on layer uncertainty, effectively reducing information loss during PartialKV inference.
Experiments conducted on two benchmarks and several models demonstrate the effectiveness of our proposed approach.

\section*{Limitations}
This paper primarily analyzes two widely-used decoder-only LMs, LLaMa~\citep{dubey2024llama} and Mistral~\citep{jiang2023mistral}. It does not include a validation study of encoder-decoder and encoder-only architectures.


\section*{Acknowledgements}
We would like to thank the anonymous reviewers and meta-reviewer for their insightful suggestions.
This work was supported by the National Natural Science Foundation of China under Grant U23B2055 and 62276077, and Shenzhen Science and Technology Program under Grant ZDSYS20230626091203008.
\bibliography{zigzagkv}

\end{document}